\documentclass[conference]{IEEEtran}
\IEEEoverridecommandlockouts

\usepackage{cite}
\usepackage{amsmath,amssymb,amsfonts}
\usepackage{graphicx}
\usepackage{textcomp}
\usepackage{xcolor}
\usepackage{algorithm}
\usepackage{algorithmicx}
\usepackage{algpseudocode}
\usepackage{float}
\usepackage{multirow}
\usepackage{subfigure}
\usepackage{float}
\usepackage{hyperref}
\usepackage{makecell}
\usepackage{color}
\usepackage{array}
\usepackage{enumitem}
\usepackage{graphicx}
\usepackage{amsmath} 
\graphicspath{ {../figures} }

\newtheorem{definition}{Definition}

\floatname{algorithm}{Algorithm}

\def\BibTeX{{\rm B\kern-.05em{\sc i\kern-.025em b}\kern-.08em
    T\kern-.1667em\lower.7ex\hbox{E}\kern-.125emX}}
\begin{document}
\title{LightTR: A Lightweight Framework for Federated Trajectory Recovery}

  

\author{\IEEEauthorblockN{Ziqiao Liu$^{1, *}$, Hao Miao$^{2, *}$\thanks{* Ziqiao Liu and Hao Miao contribute equally to this work.}, Yan Zhao$^{2, \text{§}}$\thanks{§ Yan Zhao and Kai Zheng are corresponding authors.}, Chenxi Liu$^3$, Kai Zheng$^{1, \text{§}}$, Huan Li$^4$}

\thanks{The corresponding author, Kai Zheng, is with Shenzhen Institute for Advanced Study, University of Electronic Science and Technology of China, Shenzhen, China.}
 
\IEEEauthorblockA{%
$^{1}$University of Electronic Science and Technology of China, China \\
$^{2}$Aalborg University, Denmark\\
$^{3}$Nanyang Technological University, Singapore
$^{4}$Zhejiang University, China\\
{liuziqiao@std.uestc.edu.cn, \{haom, yanz\}@cs.aau.dk, chenxi.liu@ntu.edu.sg, zhengkai@uestc.edu.cn, lihuan.cs@zju.edu.cn}
}
}
\maketitle
\begin{abstract}
With the proliferation of GPS-equipped edge devices, huge trajectory data is generated and accumulated in various domains, motivating a variety of urban applications. Due to the limited acquisition capabilities of edge devices, a lot of trajectories are recorded at a low sampling rate, which may lead to the effectiveness drop of urban applications. We aim to recover a high-sampled trajectory based on the low-sampled trajectory in free space, i.e., without road network information, to enhance the usability of trajectory data and support urban applications more effectively. Recent proposals targeting trajectory recovery often assume that trajectories are available at a central location, which fail to handle the decentralized trajectories and hurt privacy. To bridge the gap between decentralized training and trajectory recovery, we propose a lightweight framework, LightTR, for federated trajectory recovery based on a client-server architecture, while keeping the data decentralized and private in each client/platform center (e.g., each data center of a company). Specifically, considering the limited processing capabilities of edge devices, LightTR encompasses a light local trajectory embedding module that offers improved computational efficiency without compromising its feature extraction capabilities. LightTR also features a meta-knowledge enhanced local-global training scheme to reduce communication costs between the server and clients and thus further offer efficiency improvement. Extensive experiments demonstrate the effectiveness and efficiency of the proposed framework.
\end{abstract}

\begin{IEEEkeywords}
Trajectory Recovery; Lightweight; Federated Learning;
\end{IEEEkeywords}

\section{Introduction}

Thanks to the explosive adoption and development of mobile sensing devices, a massive amount of trajectories has been collected in a decentralized fashion, which empowers a variety of trajectory-based applications~\cite{luo2013finding, DBLP:conf/pkdd/ZhangWWZMZ22, zhao2018rest, DBLP:conf/kdd/LiuSZZZ21, chen2022modeling, zheng2019reference, deng2023s2tul, xiao2021vehicle}, such as traffic prediction~\cite{r6}, destination prediction~\cite{r23}, and vehicle navigation~\cite{DBLP:conf/kdd/LiuSZZZ21}. 
Nonetheless, in practice, the collected trajectory data is often sampled at a low-sampling rate~\cite{zheng2012reducing}, called incomplete trajectories (a.k.a. low-sampling-rate trajectories), which damages the effectiveness of the above-mentioned applications due to the loss of detailed information and high uncertainty. Thus, it is important to recover missing points for incomplete trajectories called \emph{trajectory recovery} 
to enable more effective utilization of these low-sampling-rate trajectories.



Due to its significance, considerable research efforts have been made to design effective trajectory recovery models~\cite{r14, luo2013finding}. Traditionally, statistical models are employed to recover incomplete trajectories using historical trajectory data~\cite{r15, r22}, accompanied by a variety of map-matching algorithms, which match original GPS coordinates with their corresponding road segments. Some recent studies~\cite{r1, r2, r3} apply neural networks to recover trajectories by learning deep representations of the trajectories. Generally, these networks are composed of a stack of spatio-temporal (ST) blocks, aiming at learning the complex spatio-temporal dependencies of trajectories. The ST-block contains base \emph{ST-operators}, which can be further categorized into convolutional neural networks (CNN)~\cite{yang2023long}, recurrent neural networks (RNN)~\cite{r2, r13}, and attention  neural networks (Attn)~\cite{r1} based \emph{ST-operators}. However, existing methods assume that the models are trained with centralized data gathered from edge devices, which incur high collection and storage costs and fail to handle decentralized training data. Moreover, with the rising concerns of privacy protection, people may be unwilling to disclose their raw trajectories to untrusted data providers, because it is dangerous that real data can be used by a malicious third party. Thus, decentralized trajectory processing, which enables privacy protection, becomes a critical issue in current trajectory-based applications to adapt existing licensing agreements and data access restrictions~\cite{r6}.

As a result, we need a new kind of decentralized learning model that can adapt decentralized trajectory data and can learn complex spatio-temporal correlations effectively. Recently, federated learning (FL) provides a solution for training a model with decentralized data on multiple clients (e.g., mobile devices, organizations, or platform centers) promising the privacy. FL is a machine learning setting where many clients collaboratively train a model under the orchestration of a central server while keeping the training data decentralized. However, existing FL-based methods~\cite{r7, r4} do not consider the inherent spatio-temporal dependencies, which are important for effective trajectory embedding~\cite{r2}. In this study, we aim to develop a novel FL-based trajectory recovery model, which can bridge the gap between decentralized data processing and complex spatio-temporal dependency modeling. Nonetheless, it is non-trivial to develop such kind of model, due to the following challenges.


\emph{Challenge I: scalability.} Existing popular trajectory recovery methods usually suffer from poor scalability, as these deep learning based models are often large, where training and inference are often time-consuming and computationally expensive. 
This limits the scalability of the trajectory recovery model on resource-constrained edge computing devices, which play a vital role in decentralized computation. In addition, these methods may incur memory overflow in large-scale trajectory learning settings since the entire network must reside in memory during training. For example, given $N$ trajectories, each of which has $L$ points, the memory cost of Attn-based \emph{ST-operators} increases quadratically with $L$ and $N$ (see Table~\ref{operators}). Nevertheless, no customized lightening module exists in current trajectory recovery models, and simply lightening these models degrades their performance dramatically~\cite{lai2023lightcts}, which also limits the scalability of trajectory learning.

\emph{Challenge II: communication cost.} During the training of FL, certain rounds of communication exist between the central server and all participating clients. Two lines of factors, such as the limited network bandwidth and explosive participating clients, can create a communication bottleneck in an FL environment, which increases latencies and decreases practicalities.
   Statistically, the collected trajectories across different clients are usually not independent and identically distributed (Non-IID) and heterogeneous, which leads to a significant increase in communication rounds to achieve convergence and makes it difficult to obtain an optimal global model.
    Systematically, a certain number of clients involve in an FL environment, while the communication capacity of each client may differ due to significant constraints in hardware, network connection, and power. 
It is highly desirable, but also non-trivial, to develop a communication cost reduction method in federated trajectory recovery, that is capable of solving these statistical and systematical issues.

To tackle the two challenges, we provide a lightweight framework for federated trajectory recovery called LightTR based on horizontal federated learning (i.e., a server-clients architecture), where the objective is to collaboratively train models by maintaining a shared global model on a central server and utilizing all clients' data in a decentralized fashion. LightTR encompasses two major modules: 
a local trajectory preprocessing and light embedding module and a meta-knowledge enhanced local-global training module.

To avoid huge memory consumption and limited scalability (\emph{Challenge I}), we design a local lightweight trajectory embedding (LTE) model for each client. Specifically, LTE contains an embedding component and a stack of ST-blocks to learn effective spatio-temporal representations. Unlike previous studies~\cite{r13}, we formalize a lightweight \emph{ST-operator} in ST-blocks and replace the popular \emph{ST-operators} (e.g., CNN and Attn) with a pure MLP (multi-layer perceptron) architecture considering the lower space complexity (i.e., $\mathcal{O}(L+D+1)$) and time complexity (i.e., $\mathcal{O}(N\cdot(L+D))$) of MLP, where $L$ denotes the number of points in each trajectory, $D$ is the embedding size, and $N$ is the number of trajectories. Here, we use only one RNN layer combined with MLP to ensure temporal dependencies capturing.

To reduce communication cost and speed up the model convergence (\emph{Challenge II}), we propose a meta-knowledge enhanced local-global training module by means of knowledge distillation. Before federated training, we propose a Teacher model (i.e., a meta-learner) to learn local meta-knowledge for each client using a part of its local data. We consider the local lightweight trajectory embedding model as the student model. During FL, the teacher model is employed to guide the optimization of the student model, in order to learn better common features and achieve faster convergence.

The major contributions are summarized as follows.
\begin{itemize}
    \item To the best of our knowledge, this is the first study to systematically learn federated trajectory recovery on decentralized trajectories. We propose a lightweight federated framework entitled LightTR to perform decentralized trajectory recovery, which offers privacy protection of locally collected trajectories.
    \item To increase the scalability, we introduce a local trajectory preprocessing and light embedding module to capture effective spatio-temporal correlations of trajectories with a customized lightweight trajectory embedding \emph{ST-operator}.
    \item To reduce the communication cost in FL, we design a meta-knowledge enhanced local-global training module by means of knowledge distillation to achieve faster convergence and better accuracy.
    \item We report on experiments using real datasets, demonstrating the effectiveness and efficiency of the proposed LightTR framework.
\end{itemize}


The remainder of this paper is organized as follows. Section~\ref{preliminaries} covers preliminary concepts and formalizes the problem of federated trajectory recovery. Section~\ref{Analysis} analyzes the drawbacks of existing trajectory learning methods. We report the design of LightTR framework in Section~\ref{Method}, followed by the experimental study in Section~\ref{experiment}. Section~\ref{relatedwork} surveys the related work, and Section~\ref{conclusion} concludes the paper.
\section{Preliminaries}
\label{preliminaries}
We proceed to present necessary preliminaries and then define the problem addressed. Table~\ref{notation} lists the notations used throughout the paper.

\begin{table}[t]
\renewcommand\arraystretch{1.1}
\footnotesize
\centering
\caption{\textbf{Summary of Notations}}
\setlength\tabcolsep{17pt}
\scalebox{1}{
\begin{tabular}{cl}
    \hline
    \textbf{Symbol} &\textbf{Definition} \\
    \hline
    $G$ & Road network\\
    $V$ & Vertex set \\
    $E$ & Edge set \\
    $v_{i}$ & Road segment intersection or road end\\
    $e_{i,j}$ & Road segment from $v_i$ to $v_j$\\
    $p_i$ & GPS Point\\
    
    $\tau$ & Raw incomplete trajectory\\
    $t_i$ & Timestamp of $p_i$\\
    $\epsilon$ & Sampling rate \\
    $\widetilde{p}_i$ & Map-matched point\\
    $r$ & Moving ratio\\
    $dis(a,b)$ & Distance between nodes $a$ and $b$\\
    $e.N_1$ & Start node of  $e$\\
    $e.N_2$ & End node of  $e$\\
    $e.N_{cur}$ & The current node of $e$\\
    $T $ & Map-matched trajectory \\
    $\mathcal{T}$ & Map-matched trajectory dataset\\
    $C_i$ & The $i$-$th$ client\\
    \hline
\end{tabular}}
\label{notation}
\vspace{-0.4cm}
\end{table}

\begin{definition}[Road Network]
    A road network is a directed graph $G=(V,E)$, where $V$ is a road vertex set, and $E$ is a road edge set. Each $v_i \in V$ denotes a road segment intersection or a road end, and each $e_{i, j} \in E$ denotes a directed road segment from $v_i$ to $v_j$.
\end{definition}

\begin{definition}[GPS Point]
    A GPS point can be defined as a triple $p=\langle lat,lng, \gamma \rangle$, which captures the latitude $lat$ and longitude $lng$ of the GPS position including additional information $\gamma$, e.g., address.
\end{definition}

\begin{definition}[Raw Incomplete Trajectory]
    A raw incomplete trajectory $\tau$ is a sequence of $n$ points, 
    i.e., $\tau = \langle (p_1, t_1), (p_2, t_2), \cdots, (p_n, t_n) \rangle$, where $t_i$ denotes the timestamp of $p_i$.
\end{definition}

\begin{definition}[Sampling Rate]
    A sampling rate $\epsilon$ is the time difference between two consecutively sampled points, which is usually determined by the sampling device.
\end{definition}

In practice, the sampling rate often changes near $\epsilon$ in most cases because of the inherent temporal bias of trajectory data. It is also worth noting that the collected trajectory points may not be precise due to GPS device measurement errors and GPS noises. Map matching is usually adopted to project raw points onto the road network~\cite{r2,  brakatsoulas2005map, rappos2018force, chambers2020map}. In this study, we use the map-matching method in the deep hierarchical network (DHN)~\cite{r17} to convert a sequence of raw latitude/longitude coordinates for map matching.



\begin{definition}[Map-matched $\epsilon$-Sampling-Rate Trajectory]
    A map-matched trajectory $T$ with $\epsilon$-sampling rate is a sequence of map-matched trajectory points, i.e., 
    \begin{equation}
        T= \langle (\widetilde{p}_1, t_1), \cdots, (\widetilde{p}_i, t_i), \cdots, (\widetilde{p}_n, t_n) \rangle,
    \end{equation}
    where $t_{i+1} - t_i = \epsilon$, $\widetilde{p}_i = \langle e, r \rangle$ is the map-matched (trajectory) point at timestamp $t_i$, $e$ denotes the road segment, and $r$ is the moving ratio. Next, $r$ represents the ratio between the current moving distance (from the start node of $e$ to the current node) and the length of road segment $e$, i.e., $r=\frac{dis(e.N_1,e.N_{cur})}{dis(e.N_1,e.N_{2})}$, where $dis(a,b)$ denotes the distance between nodes $a$ and $b$, $e.N_1$ denotes the start node of $e$, $e.N_{cur}$ denotes the current node of $e$, and $e.N_2$ denotes the end node of $e$. 
\end{definition}

As shown in Figure 2, each road segment $e$ covers the start node $e.N_1$ and end node $e.N_2$, and records the distance between the two points. Next, the moving ratio $r$ will be determined by the moving distance (from $e.N_{1}$ to the current node $e.N_{cur}$) and 
the total length of $e$.
For example, if the current node is located in the middle of $e_4$, we can obtain that $r= 0.5$.


\begin{definition}[Incomplete Map-matched $\epsilon$-Sampling-Rate Trajectory]
    An incomplete map-matched $\epsilon$-sampling-rate trajectory, i.e.,
    \begin{equation}
        T_{icp} = \langle (\widetilde{p}_1, t_1), \cdots, (\bar{p}_j, t_j), \cdots, (\widetilde{p}_n, t_n) \rangle,
    \end{equation}
    is composed of a sequence of map-matched points (e.g., $\widetilde{p}_i$) and missing points (e.g., $\hat{p}_j$).
\end{definition}

In the rest of the paper, we will use the terms \emph{map-matched point} (\emph{map-matched $\epsilon$-sampling-rate trajectory}) and \emph{point} (\emph{trajectory}) interchangeably when the context is clear.

\begin{figure}[ht]
    \centering
    \vspace{-0.2cm}
    \includegraphics[scale=0.4]{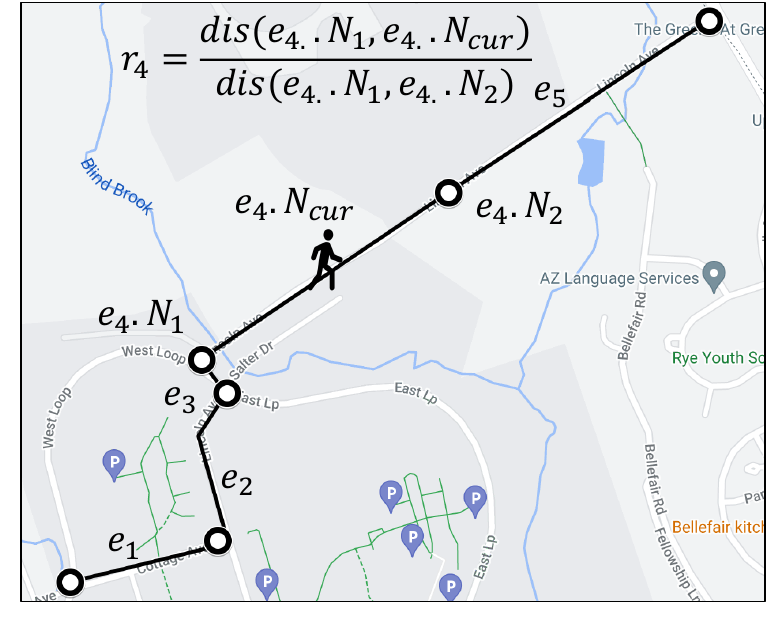}
    \vspace{-0.3cm}
    \caption{Moving Ratio}
    \vspace{-0.4cm}
    \label{fig:enter-label}
\end{figure}

\begin{definition}[Platform Center]
    A platform center $C$, e.g., the distributed data center of a company, keeps its map-matched trajectory dataset $\mathcal{T}=\{T_1, T_2, \cdots, T_l\}$, where $T_i$ is the $i$-$th$ map-matched trajectory and $l$ is the length of $\mathcal{T}$. Note that we consider a platform as a client.
\end{definition}

Based on the above definitions, we formally define the studied problem as follows.



\textbf{Federated Trajectory Recovery Problem Statement.} Given a server $\mathcal{S}$ and $N$ clients with their local trajectory datasets $\mathcal{T}= \left\{\mathcal{T}_1, \mathcal{T}_2, 
\cdots, \mathcal{T}_{N}\right\}$, each dataset $\mathcal{T}_i$ ($1\le i\le N$) in client $C_i$ is a set of incomplete map-matched trajectories
, i.e., $\mathcal{T}_i = \left\{T_{icp,1}^i, \cdots, T_{icp, l}^i\right\}$. Our problem aims to learn a shared global function $F(\cdot)$, such that for any incomplete map-matched trajectory, their missing coordinates are recovered. Formally, for each incomplete trajectory $T_{icp}$ in a platform center, we have


\begin{equation}
    \overbrace{\langle \cdots, (\bar{p}_i, t_i), \cdots, (\widetilde{p}_j, t_j), \cdots \rangle}^{\text{Incomplete Trajectory}} \stackrel{F(\cdot)}{\longrightarrow}\overbrace{\langle (\cdots, (\hat{p}_i, t_i), \cdots, (\widetilde{p}_j, t_j), \cdots \rangle}^{\text{Recovered Trajectory}},
\end{equation}
where $1 \leq i < j \leq n$.

\section{Analysis and Observations}
\label{Analysis}
In this section, we first analyze existing trajectory modeling proposals in a generic framework and investigate the complexity of representative models. Next, we identify directions for achieving a decentralized lightweight trajectory recovery framework.

\begin{figure}[!htbp] \centering
	\vspace{-0.4cm}
	\subfigure[] {
		\includegraphics[scale=0.498]{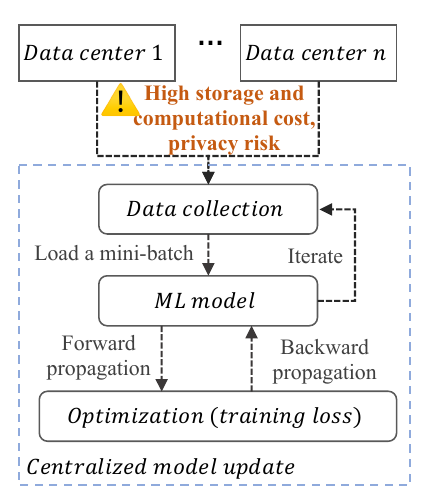} \label{centralized}
	}     
	\subfigure[]{ 
		\includegraphics[scale=0.498]{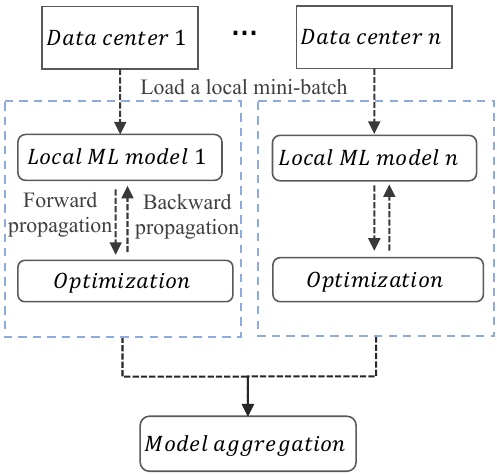}   \label{decentralized}  
	}  
	\vspace{-0.29cm}
	\caption{(a) A generic framework of existing trajectory recovery methods. (b) An illustration of the federated learning framework.}
	\label{analysis} 
	\vspace{-0.4cm}
\end{figure}

To gain insight into the prospects of the decentralized lightening trajectory modeling methods, we consider representative models~\cite{r14, r2, r1, r13, r3, luo2013finding, chen2011discovering, su2013calibrating, su2015calibrating, lun2023resisting}. Figure~\ref{analysis}(a) shows a generic framework for these models. Generally, a trajectory recovery model has three components: 
\begin{enumerate}
    \item \emph{a data collection module} that gathers data collected from different data centers into a central server;
    \item \emph{a machine learning (ML) model} consisting of stacked spatio-temporal blocks to extract spatio-temporal correlations as high-dimensional features;
    \item \emph{an optimization module} that iteratively updates the model with loaded mini-batch data by means of forward and backward propagation.
\end{enumerate}

\emph{We observe that these trajectory recovery models are trained in a centralized manner while fail to consider the decentralized setting.} 
Although models learned on such centralized trajectory data hold the promise of greatly improving usability, there are risks and responsibilities to storing trajectories in a centralized location due to their sensitive nature, which also leads to high storage and computational cost~\cite{r4}.
In addition, it is dangerous that real data can be used by a malicious third party. Thus, people may be unwilling to disclose their raw trajectories, which are especially sensitive and private. 

\emph{This calls for a decentralized trajectory learning model} where the data and computation are distributed among multiple data centers, which brings the benefit of privacy protection and potentially more efficient processing. We thus propose to design a new federated trajectory recovery framework, to be detailed in Section~\ref{FL}. Figure~\ref{analysis}(b) shows the typical learning process of a federated learning framework. Specifically, each data center trains a local model using its private data, while the central server aggregates the parameters of the learned local models periodically to collaboratively obtain a global model.

In addition, we discuss the benefits in privacy protection of the proposed framework LightTR briefly. First, LightTR alleviates sensitive information leakage since it keeps the training data decentralized and only sends model parameters to the central server. Further, LightTR mitigates malicious attacks by distributing the learning process across clients, making it harder for an untrusted third party to use the real trajectory data.

In this study, we only consider deep learning (DL) based trajectory modeling solutions, which usually are composed of a stack of spatio-temporal (ST) blocks. ST-blocks, including stacked \emph{ST-operators}, are the basic ingredients for extracting comprehensive features. We categorize popular \emph{ST-operators} into different families based on the base operators that they extend: \emph{convolutional neural network (CNN) based ST-operators, recurrent neural network (RNN) based ST-operators, and attention (Attn) based ST-operators}. A line of studies~\cite{yang2023long, fang2023heterogeneous} applies \emph{CNN-based ST-operators}, i.e., temporal convolutional networks, that applies dilated causal convolutions to trajectory data for feature extraction. Another line of research~\cite{r2, r13} uses \emph{RNN-based ST-operators} that process trajectories based on a recursive mechanism, which is usually combined with graph neural network. The rest of studies~\cite{r1, r3} adopts \emph{Attn-based ST-operators} that enables weighted temporal information extraction over long sequences by applying the attention mechanism to establish self-interactions of input trajectories. The time and space complexity of the base operators is reported in Table~\ref{operators}, where $D$ is the number of embedding dimensions, $N$ is the number of trajectories, and $L$ is the maximum length of historical trajectories (each historical trajectory may have different length).

\begin{table}
    \centering
    \footnotesize
    \setlength\tabcolsep{4pt}
    \renewcommand\arraystretch{1.6}
    \caption{Categorization and Analysis of Base \emph{ST-operators}.}
    \vspace{-0.3cm}
    \begin{tabular}{ccccccc}
    \hline
        \multicolumn{2}{c}{\textbf{Categorization}} & \textbf{Time Complexity} & \textbf{Space Complexity}  \\
        \hline
         \multirow{3}{*}{\rotatebox{90}{\emph{ST-operator}}} & CNN~\cite{yang2023long, fang2023heterogeneous} & $\mathcal{O}(D^2 \cdot N \cdot L)$ & $\mathcal{O}(D^2)$\\
         & RNN~\cite{r2, r13} & $\mathcal{O}(D^2 \cdot N \cdot L)$ & $\mathcal{O}(D^2)$\\
         & Attn~\cite{r1, r3} & $\mathcal{O}(D^2 \cdot N \cdot L \cdot (D+L))$ & $\mathcal{O}(D^2)$\\
                                                            
        \hline
    \end{tabular}
    \vspace{-0.5cm}
    \label{operators}
\end{table}

\emph{As shown in Table~\ref{operators}, we observe that ST-operators, which are the main components of trajectory learning models, incur expensive computational and storage costs}, where the time and space complexities are both proportional to $D^2$. \emph{It motivates us to develop lightweight \emph{ST-operators} for trajectory recovery.} Although it is straightforward for achieving lightness to manipulate the embedding size $D$ of \emph{ST-operators}, many studies~\cite{chen2020multi, wu2021autocts, wu2023autocts+} have shown that reducing $D$ will inevitably degrade learning capabilities of \emph{ST-operators}. 

Recent studies~\cite{liang2023mmmlp, li2023automlp} show the superior capabilities of multi-layer perceptron (MLP) (also called Dense) based architectures in feature extraction. MLP-based architectures benefit in low computational cost and fewer parameters while keeping comparable performance to mainstream \emph{ST-operators}. Thus, we propose to simplify and reduce the model computations by replacing a majority of the base network (e.g., CNN) in \emph{ST-operators} with MLP. 

\section{Methodology}
\label{Method}
\begin{figure*}[htp]
    \centering
    \includegraphics[scale=0.53]{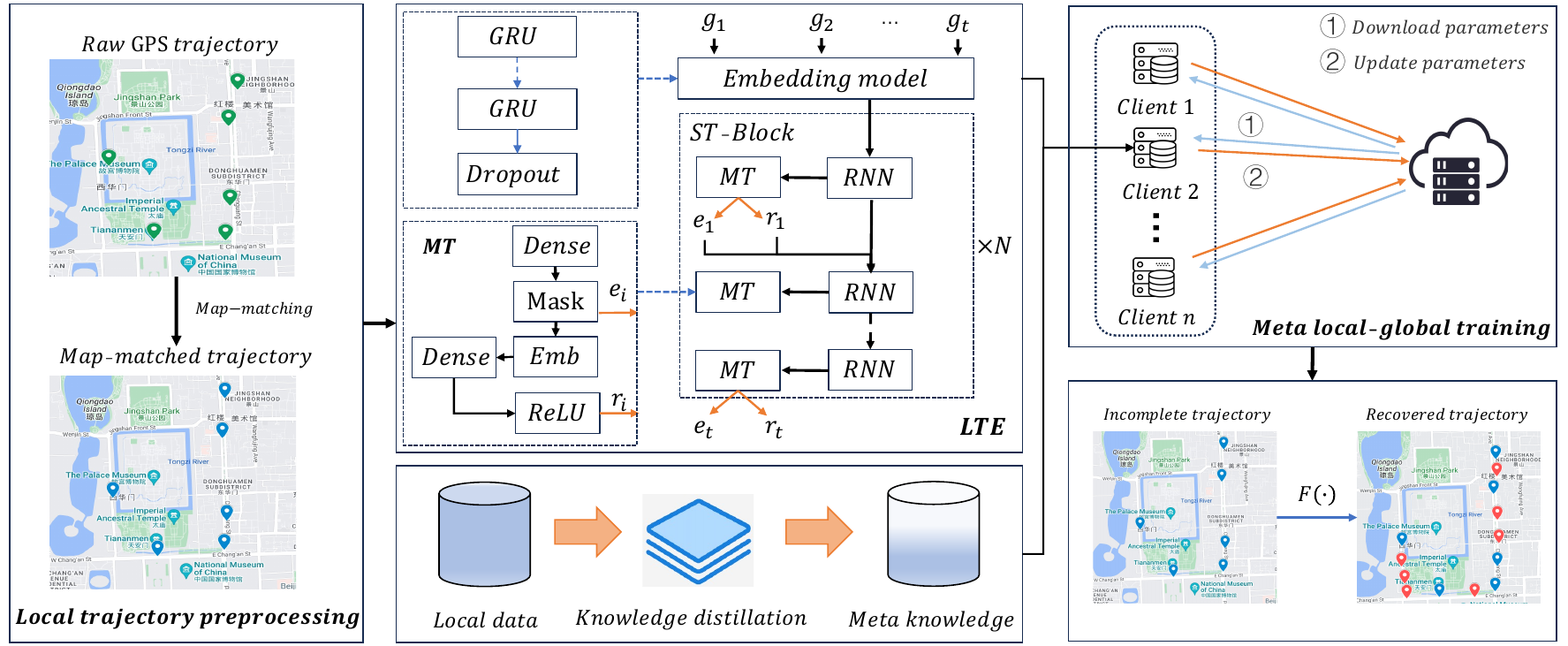}
    \caption{LightTR: Lightweight Trajectory Recovery Framework}
    \label{fig:framework}
    \vspace{-0.5cm}
\end{figure*}

In this section, we propose a federated trajectory recovery framework, entitled LightTR, as shown in Figure~\ref{fig:framework}. We first give an overview of the framework and then provide specifics on each module in the framework.
\subsection{Framework Overview}

As illustrated in Figure~\ref{fig:framework}, LightTR is based on horizontal federated learning, i.e., a client-server architecture, to collaboratively train local trajectory learning models under the orchestration of a central server, while keeping the local data private. LightTR can be divided into two major modules: local trajectory preprocessing and light embedding as well as meta-knowledge enhanced local-global training.

\begin{itemize}

\item \emph{Local Trajectory Preprocessing and Light Embedding.} In this module, each client first preprocesses the collected trajectory data to generate the map-matched trajectories by DHN~\cite{r17}. Moreover, we design a lightweight local trajectory embedding module aiming at reducing the computation cost. Finally, each client will download parameters and learn a local model to recover trajectories.


\item \emph{Meta-knowledge Enhanced Local-Global Training.} In this module, the central server will aggregate the updated parameters from selected clients to obtain a global model. In particular, a novel meta-knowledge enhanced local-global training module is proposed to accelerate the model convergence and reduce the communication cost between clients and the central server by means of knowledge distillation.
\end{itemize}

Next, we will provide the technical details of each module, respectively.
\subsection{Local Trajectory Preprocessing and Light Embedding}
\label{FL}

In this section, we report the preprocessing of local trajectories and the proposed lightweight trajectory embedding module in a client. Note that the model for the central server shares the same trajectory learning model architecture, which will be elaborated in Section~\ref{local_model}, with models for all clients.

\subsubsection{Local Trajectory Preprocessing}

Given a road network, we first map all collected trajectories into the corresponding road intersections by converting their GPS location into discrete units, referring to the Hidden Markov Model (HMM) in DHN~\cite{r17}. Specifically, given a low-sampling-rate trajectory $\tau$, we obtain the corresponding map-matched trajectory $T$ as follows: 
\begin{equation}
\begin{split}
    \tau &= \langle(p_1, t_1), \cdots, (p_i, t_i), \cdots (p_n, t_n)\rangle\\
    T &= \langle(g_1, t_1), \cdots, (g_i, t_i), \cdots (g_n, t_n)\rangle = HMM(\tau),
\end{split}
\end{equation}
where $g_i = (x_i, y_i, tid_i)$, $\forall 1 \leq i \geq n$ is the converted unit, and $x_i$ and $y_i$ represents the $i$-$th$ grid cell. We extract $tid_i = \lfloor\frac{t_i - t_0}{\epsilon} \rfloor$ to guide the model to learn how many points should be recovered between two consecutive low-sampling-rate points.

\subsubsection{Lightweight Trajectory Embedding}
\label{local_model}
As shown in Figure~\ref{fig:framework}, Lightweight Trajectory Embedding (LTE) module is composed of an embedding model and stacked ST-blocks. The embedding model encodes the map-matched trajectory $T$ into a single vector to capture its complex sequential dependencies, while the ST-blocks, containing a customized lightweight \emph{ST-operator}, aim to predict the road segment $e$ and moving ratio $r$ of missing points, simultaneously.

We proceed to elaborate the \emph{embedding model} and the lightweight \emph{ST-blocks}.

\textbf{Embedding Model.}
Given a map-matched incomplete trajectory $T_{icp} = ⟨(\widetilde{p_1},t_1), \cdots, (\widetilde{p_i},t_i), \cdots, (\widetilde{p_n},t_n)⟩$, the embedding model aims to convert $T$ into a single vector to capture the complex spatial and temporal correlations. 
It is important to model the sequential dependencies of trajectories, which enables accuracy~\cite{r3}. We adopt Gated Recurrent Unit (GRU) as the embedding layer since GRU brings the benefits of long-term temporal dependency capturing without performance decay and efficient computation. We input the map-matched raw trajectory $T$ into GRU to obtain its hidden features, which are then input into the ST-blocks. Specifically,  we pass the input trajectory $T$ along the time span $[1, 2, \cdots, n]$ through the GRU. For each time step $t$, the hidden features can be formulated as:
\begin{equation}
\begin{split}
    r_t &= \sigma(W_r \cdot [h_{t-1},g_{t}] + b_r)\\
    z_t &= \sigma(W_z \cdot [h_{t-1},g_{t}] + b_z)\\
    \tilde{h_t} &= tanh(W_{h} \cdot [r_i \ast h_{t-1},g_{t}] + b_h)\\
    h_t &= (1-z_t) * h_{t-1} + z_i * \tilde{h_t},
\end{split}
\end{equation}
where $W$ represents the weight for respective gate neurons and $b$ is the bias for the respective gate, and $[\cdot]$ denotes the feature concatenation.

For simplicity, the embedding model derives the hidden features $h_t$ of the encoder for low-sampling-rate trajectories embedding:
\begin{equation}
    h_t = embedding(h_{t-1}, s_{t-1}),
\end{equation}
where the last state $h_t$ is considered as the learned hidden features, which is also used as the input of ST-blocks.

\textbf{ST-blocks.}
The learned hidden features $h_t$ are then input into the stacked ST-blocks to obtain the data representations for incomplete trajectory recovery. As illustrated in Figure~\ref{fig:framework}, to increase scalability and reduce computation cost, we design a lightweight \emph{ST-operator} including an RNN layer where each RNN cell is followed by a pure MLP-based multi-task (MT) model, which benefits from the low space and time complexities of MLP. RNN can only predict the numerical coordinates of trajectories, while cannot ensure the prediction of missing points being map-matched onto the road network. Thus, the lightweight \emph{ST-operator} contains a novel MT model to predict the road segment $e$ and moving ratio $r$ at the same time due to their high correlations, leveraging multi-task learning~\cite{pan2009survey}.

More specifically, we feed $h_t$ into RNN to obtain high-dimensional hidden features $h_t^\prime$, which capture the sequential dependencies. Next, $h_t^\prime$ is input into the MT model to predict the road segment $e_t$ and moving ratio $r_t$ simultaneously. The lightweight \emph{ST-operator} can be formulated as:
\begin{equation}
    \begin{split}
        &h_t^\prime = RNN(h_t)\\
        &e_t, r_t = MT(h_t^\prime),
    \end{split}
\end{equation}


The left center in Figure \ref{fig:framework} shows detailed architecture of the MT model to predict $e_t$ and $r_t$, simultaneously. Specifically, after obtaining hidden features through RNN, we first use a dense (MLP) layer with a constraint mask layer (Mask) and a road segment embedding layer (Emb) to predict the road segment $e_t$, which can be formulated as follows.
\begin{equation}
\begin{split}
    &h_{t, d} = Dense(h_t^\prime, W_d) = W_d \cdot h_t^\prime + b_d\\
    &e^t = Mask(h_{t, d})\\
    &h_{t, e} = Emb(h_{t, d}, e^t) = ReLU(h_{t, d}+RNN(e^t))\\
    &r_t = ReLU(Dense([h_{t, e}, e^t], W_r)) = W_r \cdot [h_{t, e}, e^t] + b_r,
\end{split}
\end{equation}
where $W$ and $b$ denote the trainable parameters and bias. 


Then, $e_t$ together with $h_t^\prime$ are concatenated to go through the dense layer with an activation function (i.e., ReLU) to predict moving ratio $r_t$. Next, $e_t$ and $r_t$ affect $e_{t+1}$ and $r_{t+1}$. For example, people may be more willing to visit places that are close to their current locations. We also input $e_t$ and $r_t$ into RNN. When inferring the missing points between two consecutive points of a low-sampling-rate trajectory, the hidden state $h_t$ of our ST-blocks can be represented as: 
\begin{equation}
    h_t = STBlocks(h_{t-1},e_{t-1},r_{t-1}),
\end{equation}
where $e_{t-1}$ and $r_{t-1}$ represent the road segment embedding of the predicted road segment and its moving ratio at timestamp $t-1$, respectively.




\textbf{Constraint Mask Layer.}
In order to reduce the training complexities and achieve fine-grained trajectory recovery, we employ a constraint mask layer~\cite{r2}. 
We use an exponential function (shown in Eq.~\ref{omega}) to capture the influence of distance.
\begin{equation}
\label{omega}
    c_i = exp(\frac{-dist^2(p_i, \widetilde{p_i})}{\gamma}),
\end{equation}
 where $dist(p_i, \widetilde{p_i})$ denotes the Euclidean distance between the original point $p_i$ and its map-matched point $\widetilde{p_i}$ on corresponding road segment $e$, $\gamma$ is a parameter related to the road network ($\gamma = 125$ in our work). Note that we only consider points existing in low-sampling-rate trajectories that are not far away from road segments. If the distance between a road segment $e$ and point $p_i$, we set $\omega(e,p_i)$ as 0. Finally, we combine Eq.~\ref{omega} with softmax as the constraint mask layer, which can be defined as follows.
\begin{equation}
    P(e_i|h_i) = \frac{exp(h_{t,d}^T \cdot w_c) \odot c_i}{\sum_{c'\in C}exp(h_{t,d}^T \cdot w_{c'})\odot c_i},
\end{equation}
where $w_c$ is a trainable parameter matrix. We finally use $argmax$ to get the final prediction of road segment $e^t$.





\subsection{Meta-knowledge Enhanced Local-Global Training}

\begin{figure}[t]
    \centering
    \includegraphics[scale=0.29]{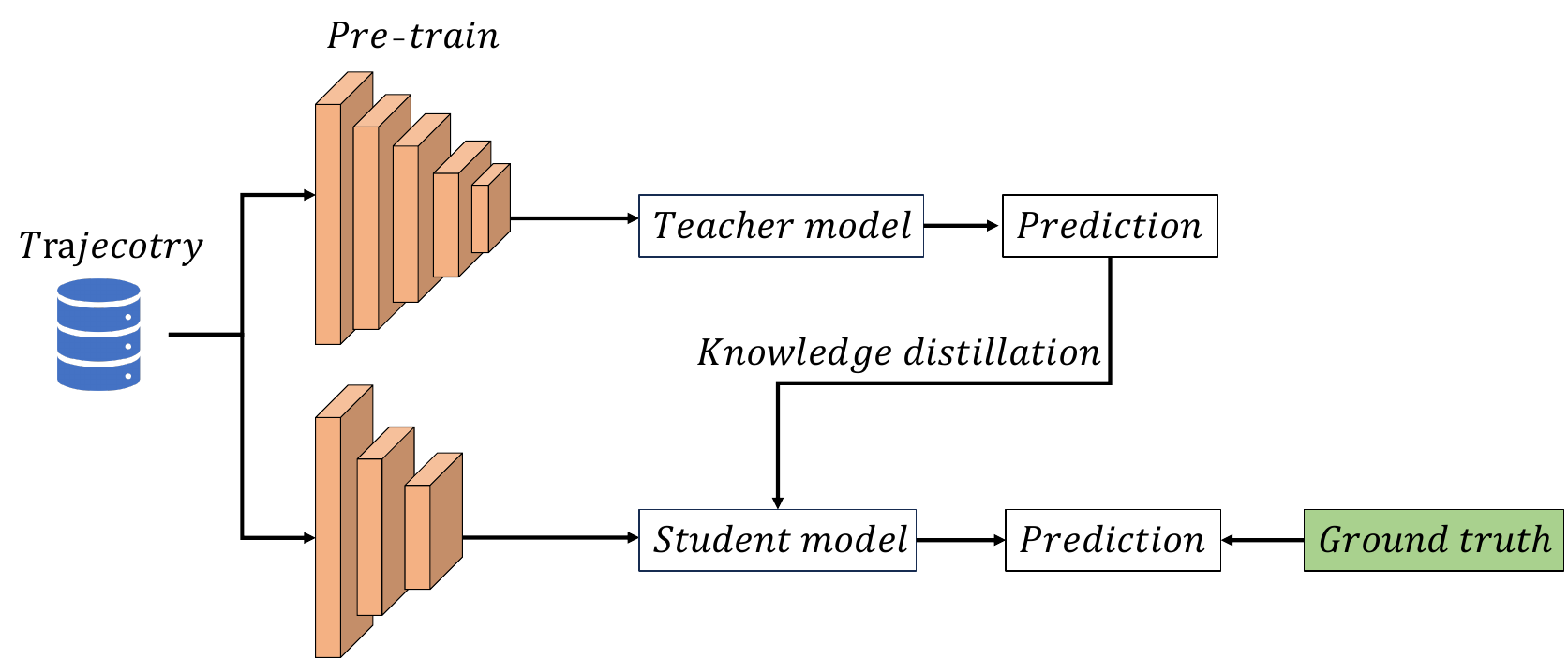}
    \vspace{-0.2cm}
    \caption{The Process of Knowledge Distillation}
    \label{KD}
    \vspace{-0.6cm}
\end{figure}

To reduce the communication cost, we propose a meta-knowledge enhanced local-global training (or meta local-global training for short) module for model aggregation by means of knowledge distillation. Figure~\ref{KD} shows the process of knowledge distillation, which includes a teacher model and a student model. Specifically, we first pre-train a teacher model, also called a meta-learner, to learn meta-knowledge for each client with a subset of local trajectories. During local training, each client first downloads parameters from the central server, and the learned meta-knowledge is used to guide the local model (i.e., lightweight trajectory embedding) learning. We consider the local model as the student model. Finally, the central server aggregates parameters to update the global model with a subset of clients. In this study, we use the lightweight trajectory embedding model as the student and teacher model.


In federated trajectory recovery, we have $N$ clients, denoted as $\{C_1, C_2, \cdots, C_N\}$, with their own datasets $\{\mathcal{T}_1, \mathcal{T}_2, \cdots, \mathcal{T}_N\}$. Each dataset $\mathcal{T}_i$ consists of three parts, a training, validation, and test dataset, denoted as $\mathcal{T}_i^{train}$, $\mathcal{T}_i^{valid}$, and $\mathcal{T}_i^{test}$, respectively. We aim to learn a good trajectory recovery model $f_i(\cdot)$ for each client on its local dataset $\mathcal{T}_i$:
\begin{equation}
    min \frac{1}{N} \sum_{i=1}^N \frac{1}{n_i^{test}} \sum_{j=1}^{n_i^{test}}L_{total}(f_i(T_{icp}^j), T_{ground}^j),
\end{equation}
where $L_{total}$ is the overall loss function, $n_i^{test}$ is the length of $\mathcal{T}_i^{test}$, and $T_{ground}$ is the ground truth of complete trajectory.

We proceed to introduce the local trajectory recovery objective function and meta-knowledge enhanced local-global training.

\subsubsection{Local Trajectory Recovery Objective} 
As we aim to predict road segment and moving ratio simultaneously, we combine the cross-entropy loss $L_1(\theta)$ and mean squared error $L_2(\theta)$ as the local loss function $L_{local}(\theta)$ for trajectory recovery. $L_1(\theta)$ is used for road segment prediction, while we use $L_2(\theta)$ for moving ratio prediction.
\begin{equation}
\label{localloss}
    L_{local}(\theta) = L_1(\theta) + \mu L_2(\theta),
\end{equation}
where $\mu$ is a tunable parameter to balance the trade-off linearly between the road segment and moving ratio prediction.

The cross-entropy loss is formulated as follows.
\begin{equation}
\begin{split}
    L_1(\theta) = - \sum_{(T,\tilde{T}\epsilon \mathcal{T})} &\sum_{j=1}^{|\tilde{T|}} \sum_{l=1}^{L}1{a_j\cdot e_l}log(R_\theta(\hat{a}_{j} e = e_l | d_1:j-1))\\
     &s.t. d_{j-1} = (T,\tilde{T}_{1:j-1}),
\end{split}
\end{equation}
where $T$ and $\tilde{T}$ represent incompleted trajectories and recovered completed trajectories, respectively, $L$ is the number of road segments, $a_je$ is the ground truth of road segment, $\hat{a}_j$ is the prediction, and $R_\theta$ represents the lightweight trajectory embedding module. Next, $D$ is the training data.

The mean squared error is also employed to predict the movement ratio, which can be formulated as:
\begin{equation}
\begin{split}
    &L_2(\theta) = - \sum_{(T,\tilde{T}\epsilon \mathcal{T})} \sum_{j=1}^{|\tilde{T|}}(a_jr - R_\theta(d_{j-1})^2)\\ 
    &s.t. d_{j-1} = (T,\tilde{T}_{1:j-1}),
\end{split}
\end{equation}
where $a_jr$ is the real moving ratio.

\subsubsection{Meta-knowledge Enhanced Local-global Training.}
To achieve faster convergence and reduce the communication cost, we propose a meta-knowledge enhanced local-global training module by means of knowledge distillation. We first learn a meta-learner (a teacher model) to accumulate meta-knowledge for each client. Each client trains a local model after downloading parameters from the server with meta-knowledge guiding the local training. The central server obtains the global model by aggregating parameters from the participating clients.


We employ the proposed lightweight trajectory embedding model as the teacher model $f_{tea}$ to perform trajectory missing points prediction task with incomplete trajectories. The teacher model is then used to guide the local model $f_{stu}$ training via knowledge distillation. The knowledge distillation can be formulated as:
\begin{equation}
    L_{dist}(f_{tea},f_{stu};T_{icp}) = ||f_{tea}(T_{icp})-f_{stu}(T_{icp})||_2^2,
\end{equation}
where $||\cdot||_2$ represents $L2$-$norm$.



Through knowledge distillation, we can make good use of meta-knowledge to guide the local training for each client. Therefore, the total loss $L_{total}$ function for each client $C_i$ is:
\begin{equation}
\begin{split}
\label{totalloss}
    L_{total} = \frac{1}{|\mathcal{T}^{train}|} \sum_{(T_{icp}, T_{ground})\epsilon \mathcal{T}_i^{train}} L_{local}(R_{\theta}(T_{icp}), T_{ground}) \\+ \lambda L_{dist}(f_{tea}, R_{\theta}; T_{icp}),
\end{split}
\end{equation}
where $\lambda$ is a trade-off knowledge transfer and focusing on the current data, and $L_{local}$ is the trajectory recovery loss.

\begin{algorithm}[h]
    \caption{Teacher model training}
    \label{MetaKnowledge}
    \small{
    \begin{algorithmic}[1]
    \Require subsets of Dataset $\left\{\mathcal{T}_i\right\}_{i=1}^N$ in $N$ clients, $\lambda_0$, $l_t$
    \Ensure  A common teacher model $f_{tea}$
    \State $\lambda\leftarrow\lambda_0$
    \label{s1}
    \State Train local teacher model $f_i$ using $L_{local}$ with $\mathcal{T}_i^{train}$ in each client $C_i$
    \label{s2}
    \State Send the current model $f_i$ to the next client $C_{i+1}$
    \label{s3}
    \State $acc_{i+1}^{vaild} \leftarrow $ $f_{i+1}(\mathcal{T}_{i+1}^{valid})$
    \label{s4}
    \If {$acc_{i+1}^{valid}$ > $l_t$}
    \State Train $f_{i+1}$ with $\mathcal{T}_{i+1}$ with Eq.\ref{totalloss}
    \label{s5}
    \Else
    \State $f_i \leftarrow f_{i+1}$
    \State Train $f_{i+1}$ with $\mathcal{T}_{i+1}$ with Eq.\ref{totalloss}
    \EndIf
    \label{s6}  
    \State Repeat steps 3 -- 10 until convergence
    \State $f_{tea} \leftarrow f_{N}$
    \end{algorithmic}}
\end{algorithm}

Algorithm~\ref{MetaKnowledge} specifies the process of teacher model training, which is used for all clients. Here, we train all clients sequentially in a cyclic way, and the previous knowledge is transferred to the next one. The previously learned knowledge will be preserved if it is useful for the current client training. Otherwise, it will be discarded. The data heterogeneity across clients is alleviated in this way. We fix $\lambda = \lambda_0$ to ensure preserving enough common knowledge (line~\ref{s1}). The threshold $l_t$ is used to determine whether to completely preserve the previously learned knowledge (lines~\ref{s2}--\ref{s6}). When accuracy $acc_{i+1}^{valid} < l_t$, it means too little information contained in the current client $C_{i+1}$. We directly initial the current model with the previous one to fully leverage previous knowledge, and thus handle the data heterogeneity.



\begin{algorithm}[ht]
    \caption{Meta-Knowledge Enhanced Local Training}
    \label{localmodel}
    \small{
    \begin{algorithmic}[1]
    \Require Datasets $\left\{\mathcal{T}_i\right\}_{i=1}^N$ in $N$ clients, $\lambda_0$, $l_t$, teacher model $f_{tea}$, training epochs $epoch_{max}$
    \Ensure Local Trajectory Recovery models $\left\{R_{\theta}^{i}\right\}_{i=1}^N$ 
    \State $\lambda \leftarrow 0$
    
    \label{a1}
    \label{a2}
    \State $c_{n} \leftarrow 0$

    \While{$c_{n} < N$}
    \label{a3}

    \State $acc^{train}_{i} \leftarrow R_\theta^{i}(\mathcal{T}_{i}^{train})$
    \State Train $R_\theta^{i}$ with Eq.~\ref{totalloss}
    \label{a4}
    \State $acc_{i, tea}^{valid} \leftarrow f_{tea}(\mathcal{T}_{i}^{valid})$
    \State $acc_{i}^{valid} \leftarrow R_\theta^{i}(\mathcal{T}_{i}^{valid})$
    \label{a5}
    \If{$acc_{i, tea}^{valid} \leq acc_{i}^{valid}$ and $acc_{i}^{valid}<l_t$}
    \State $\lambda \leftarrow 0$
    \label{a6}
    \Else 
    \State $\lambda \xleftarrow{} \lambda_0 \cdot 10^{min(1,(acc_{i, tea}^{valid}-acc_{i}^{valid})*5)-1}$
    \label{a7}
    \EndIf
    \EndWhile
    \State Repeat steps~\ref{a3} -- ~\ref{a8} in $epoch_{max}$ iterations
    \State \textbf{Return} $\left\{R_{\theta}^{i}\right\}_{i=1}^N$
    \label{a8}
    \end{algorithmic}
    }
\end{algorithm}

Algorithm~\ref{localmodel} illustrates the process of meta-knowledge enhanced local training (i.e., knowledge distillation) to train a good local trajectory recovery model for upcoming central aggregation. Specifically, we first set $\lambda$ to 0, which means no knowledge learned from the teacher model (line~\ref{a1}). For each client, we apply gradient descent to update its local model $R_{\theta_i}$ with local training data with Eq.~\ref{totalloss} (lines~\ref{a3}--\ref{a4}). When accuracy $acc_{i, tea}^{valid}$ of the teacher model is lower than $acc_{i}^{valid}$ of the current local model on local validation data, we keep $\lambda$ as 0, meaning that no guidance from the teacher model as the teacher model has less knowledge of the current client, and we want to refer little on it. Otherwise, we dynamically update $\lambda$ as follows:
\begin{equation}
\label{lambdaupdate}
    \lambda \xleftarrow{} \lambda_0 \cdot 10^{min(1,(acc_{i, tea}^{valid}-acc_{i}^{valid})*5)-1},
    \vspace{-0.1cm}
\end{equation}
Eq.~\ref{lambdaupdate} shows that the better the teacher model’s performance is, the larger the value of $\lambda$ is, which means that the local model learns more knowledge from the teacher model. After several iterations of model updating, we obtain $N$ local models for the preceding central aggregation (lines~\ref{a4}--\ref{a8}).

Algorithm~\ref{Updating} shows the process of local-global parameter updating. Especially, at each communication round $r$, we randomly select a subset of $N$ clients $C$, which further reduces the communication cost. The central server transfer its current model $\theta_s^{r-1}$ to selected clients (lines~\ref{b1}--\ref{b3}). Each client performs meta-knowledge enhanced local training according to Algorithm~\ref{localmodel}, which optimizes a local empirical risk objective, i.e., Eq.~\ref{totalloss} (lines~\ref{aaa}--\ref{b5}). Next, the selected clients upload their local model parameters $\theta_{c_i}$  to the central server. Finally, the central server aggregates the uploaded parameters to update the global model $\theta_s$ until it converges to a stationary point $\theta_s$ (line~\ref{b6}).

\begin{algorithm}[h]
    \caption{Local-global Parameter Updating}
    \label{Updating}
    \small{
    \begin{algorithmic}[1]
    \Require communication rounds $R$, initial global model of server $\theta_s^0$, number of clients $N$, learning rate $\alpha$, local training epochs $E$
    \Ensure final global model of server $\theta_s$
    \For{each round $r$ in $[1,2,\cdots,R]$}
    \label{b1}
    \State Randomly select $C \in [1, N]$ clients from $N$ clients
    \label{b2}
    \For{each client $c_i \in C$}
    \State Server transmits $\theta_s^{r-1}$ to client $c_i$
    \label{b3}
    \For{each local epoch $e$ in $[1,2,\cdots,E]$}
    \label{aaa}
    \State $\theta_{c_i}^r \leftarrow argmin_{\theta}L_{total}(\theta_{c_i}^r)$ 
    \State $\bigtriangledown L_{total}(\theta_{c_i}^r) \leftarrow \bigtriangledown L_{total}(\theta_{c_i}^{r-1}) - \alpha(\theta^r_{c_i}-\theta^{r-1}_{c_i})$
    \label{b4}
    \EndFor
    \State $\theta_{c_i} \leftarrow \theta_{c_i}^r$
    \EndFor
    \label{b5}
    \State $\theta_s \leftarrow \sum_{c_{i}\in C} \frac{1}{C}\theta_{c_i}$ 
    \label{b6}
    \EndFor
    \end{algorithmic}
    }
\end{algorithm}
\section{Experiment}
\label{experiment}

\begin{table}
\renewcommand\arraystretch{1.0}
\footnotesize
\centering
\caption{\textit{Statistics of Datasets}}
\vspace{-0.3cm}
\setlength\tabcolsep{6pt}
\scalebox{1}{
\begin{tabular}{ccc}
    \hline
    \textbf{Dataset} & Geolife &Tdrive\\
    \hline
    City name &Beijing &Beijing\\
    Time span & Apr 2007 to Aug 2012 & Feb 2, 2008 to Feb 8, 2008\\
    Taxi driver &182 &10357\\
    \hline
    &\textbf{Trajectory attribute}&\\
    Total length &1.2 million km &9 million km\\
    Type &GPS &GPS\\
    \hline
\end{tabular}}
\label{Dataset}
\vspace{-0.4cm}
\end{table}

\begin{table*}
    \footnotesize
    \centering
    \caption{Overall Performance Comparison on Two Datasets}
    \vspace{-0.2cm}
    \renewcommand\arraystretch{1.0}
    \setlength\tabcolsep{5pt}
    \begin{tabular}{c|c|c|c|c|c|c|c|c|c|c|c|c|c}
    \hline
    \multirow{2}{*}{\textbf{Datasets}} & \multirow{2}{*}{\textbf{Baseline}} & 
    \multicolumn{4}{c|}{\textbf{6.25\%}} & \multicolumn{4}{c|}{\textbf{12.5\%}} &
    \multicolumn{4}{c}{\textbf{25\%}}\\
        \cline{3-14} & & Recall & Precision & MAE & RMSE & Recall & Precision & MAE & RMSE & Recall & Precision & MAE & RMSE \\
        \hline
        \multirow{5}{*}{\textbf{Geolife}}
        &\textbf{FC+FL} & 0.212 & 0.219 & 0.755 & 1.015& 0.283 &0.215 &0.728 & 1.007& 0.315& 0.294& 0.685& 0.807\\
        &\textbf{RNN+FL} & 0.392 & 0.417& 0.630& 0.776& 0.397& 0.394& 0.628& 0.741 & 0.456& 0.438& 0.561& 0.659\\
        &\textbf{MTrajRec+FL}&\underline{0.595}&0.601&0.462&0.634&0.599&0.623&0.455&0.636
        &0.634&0.673&0.410&0.546\\
        &\textbf{RNTrajRec+FL} &0.594 &\underline{0.618} &\underline{0.455} &\underline{0.617} &\underline{0.631} &\underline{0.655} &\underline{0.431} &\underline{0.611} &\underline{0.669} &\underline{0.702} &\underline{0.398} &\underline{0.501}\\
        &\textbf{LightTR} & \textbf{0.674}& \textbf{0.679} & \textbf{0.375} & \textbf{0.464} &\textbf{0.724} &\textbf{0.748} &\textbf{0.335} &\textbf{0.432} &\textbf{0.758} &\textbf{0.739} &\textbf{0.323} &\textbf{0.435}\\
        \hline
        \hline
        \multirow{5}{*}{\textbf{Tdrive}}
        &\textbf{FC+FL} &0.211 & 0.198& 0.753 &1.001 &0.283 &0.232 &0.723 &1.022 &0.276 &0.270 &0.662 &0.804\\
        &\textbf{RNN+FL} &0.358 &0.355 &0.584 &0.750 &0.392 &0.396 &0.547 &0.682 &0.456 &0.464 &0.536 &0.664\\
        &\textbf{MTrajRec+FL} & 0.512&0.554&0.532&0.697&0.543&0.587&0.486&0.628
        &\underline{0.633}&0.685&0.452&0.530\\
        &\textbf{RNTrajRec+FL} &\underline{0.568} &\underline{0.611} &\underline{0.488} &\underline{0.651} &\underline{0.598} &\underline{0.643} &\underline{0.447} &\underline{0.585} &0.620 &\underline{0.705} &
        \underline{0.384} &\underline{0.473}\\
        &\textbf{LightTR} &\textbf{0.624} &\textbf{0.674} &\textbf{0.395} &\textbf{0.473} &\textbf{0.667} &\textbf{0.707} &\textbf{0.366} &\textbf{0.452} &\textbf{0.708} &\textbf{0.732} &\textbf{0.332} &\textbf{0.446}\\
        \hline
        \end{tabular}
        \vspace{-0.4cm}
        \label{baseline}
\end{table*}

\subsection{Experiment Setup}
\subsubsection{Datasets}
The experiments are carried out on two real-world public trajectory datasets: Tdrive and Geolife.
\begin{itemize}
    \item \textbf{Tdrive.} The Tdrive dataset contains taxi-trip-based trajectories in Beijing from February 2, 2008, to August 2, 2008, including 10357 taxi drivers and approximately 15 million trajectory points.
    \item \textbf{Geolife.} The Geolife dataset contains 17621 taxi GPS trajectories in Asia from April 2007 to August 2012, including 182 taxi drivers.
\end{itemize}

\subsubsection{Evaluation Metrics.} 
We aim to recover low-sampling-rate trajectories in free space to high-sampling-rate trajectories mapped onto the road network. Thus, both accuracy of road segments recovery and distance error of location inference are adopted to compare the performance of our model and baseline methods.

\textbf{Recall \& Precision.} 
We use recall and precision to evaluate the performance of route recovery by comparing the recovered road segments $P_R$ to the corresponding ground truths $G$. We define recall and precision as follows.
\begin{equation}
    Recall = \frac{|P_R \cap G|}{|G|},
    Precision = \frac{|P_R \cap G|}{|P_R|},
\end{equation}

\textbf{MAE \& RMSE.} 
Two distance measurements, i.e., Mean Absolute Error (MAE) and Root Mean Square Error (RMSE), are used to evaluate the point recovery performance. Note that we calculate the distance error based on the road network by updating the earth distance to road network constrained distance. The smaller MAE and RMSE are, the better performance the model presents. We define MAE and RMSE as:
\begin{equation}
\begin{split}
    MAE &= \frac{1}{m} \sum_{j=1}^m |dis(g_j,\hat{g_j})|, \\
    RMSE &= \sqrt{\frac{1}{m} \sum_{j=1}^m (|dis(g_j,\hat{g_j})|)^2} \\
    s.t. dis(g_j,\hat{g_j}) &= min(rn_{dis}(g_j,\hat{g_j}),rn_{dis}(\hat{g_j},g_j)),
\end{split}
\end{equation}
where $g_j$ is the ground truth location, $\hat{g_j}$ is the predicted map-matched trajectory point, and $rn_{dis}(g_j,\hat{g_j})$ is the distance of shortest path between prediction and ground truth. Meanwhile, we use $min(rn_{dis}(g_j,\hat{g_j}), rn_{dis}(\hat{g_j},g_{j}))$ as the final error since the road network is a directed graph, where the distance from $g_i$ to $g_j$ maybe not equal to the distance from $g_j$ to $g_i$.

\subsubsection{Baseline.} We compared the proposed LightTR with the following baselines. To our best knowledge, there is currently no FL-based method for discrete trajectory recovery. For fair comparisons, we transfer existing centralized baselines for trajectory recovery into their federated version by combining them with FedAvg~\cite{r4}.
\begin{itemize}
    \item \textbf{FC+FL.} 
    The FC+FL method combines horizontal FL with stacked fully connected layers (FC)~\cite{r40} for trajectory recovery. Here, we apply HMM as the map-matching algorithm.
    \item \textbf{RNN+FL.} 
    The RNN+FL method is a decentralized trajectory recovery model, where stacked RNNs are integrated with horizontal FL to collaboratively learn trajectory representations.
    \item \textbf{MTrajRec+FL.} 
    The MTrajRec+FC devises a decentralized trajectory recovery model, where the MTrajRec~\cite{r2} is employed as the local model. MTrajRec~\cite{r2} is the state-of-the-art trajectory method based on Seq2Seq.
    \item \textbf{RNTrajRec+FL.} 
    The RNTrajRec+FL method is a horizontal FL trajectory recovery model, where we set the local model as RNTrajRec~\cite{r30}. RNTrajRec applies graph neural network to learn rich spatio-temporal correlations of trajectories.
\end{itemize}

\subsubsection{Implementation Details.} We implement our model with the Pytorch framework on a GPU server with NVIDIA GTX 1080Ti GPU. The parameters in the model are set as follows. We set training epochs to 50 for each client. The initial learning rate is 0.001. The hidden features dimension in the local model is set as 512. Meanwhile, to prevent overfitting, we set the dropout ratio as 0.5 in the embedding module. In addition, the number of clients is set to 20 as default and each client owns its decentralized data. The code and additional materials (e.g., training convergence curve) can be found at \url{https://github.com/uestc-liuzq/LightTR}.

\subsubsection{Data Preprocessing}
We split the datasets into training, validation, and testing sets with a splitting ratio of 7:2:1. Since the datasets are completely sampled, we randomly remove points to transform high-sampling-rate trajectories into low-sampling-rate trajectories with a keep ratio. In this study, we set the keep ratio as 6.25\%, 12.5\%, and 25\%. Six points between each two consecutive points in an incomplete trajectory are required to be restored averagely. Referring to~\cite{r2}, we apply HMM on original trajectories to get ground truth for subsequent model performance comparison.

\subsection{Experiment Results}
\subsubsection{Overall Performance Comparison.} 
Table~\ref{baseline} shows the performance comparison among different methods on the two datasets. The best performance by a baseline method is underlined, and the overall best performance is marked in bold. The observations are as follows.
\begin{itemize}
    \item 
    Our LightTR achieves the best results among all the baselines on the two datasets with different settings of keep ratio (i.e., 6.25\%, 12.5\%, and 25\%). LightTR performs better than the best among the baselines by up to 14.7\% and 13.19\% with keep ratio=12.5\% for Recall and Precision, respectively, while obtaining MAE and RMSE reduction by at most 22.3\% and 29.2\%, on Geolife. This is because of the powerful feature capability of the proposed light \emph{ST-operator}. Moreover, the learned meta-knowledge alleviates the data heterogeneity across clients, which further improves the model performance. We also observe that the performance improvements obtained by LightTR on Geolife exceed those on Tdrive, due to the fact that the Geolife data are more sufficient than Tdrive. Thus, the methods trained with more training data lead to better results.
    \item 
    All RNN-based methods (i.e., RNN+FL, MTrajRec+FL, and RNTrajRec+FL), which is able to learn temporal dependencies of trajectories, perform better than FC+FL on both Tdrive and Geolife. It shows the importance of effective temporal dependency capturing for trajectory embedding. FC+FL performs the worst on both datasets, which demonstrates that simply stacked FC layers are too shallow to capture the complex temporal correlations and thus shows the necessity of the newly proposed lightweight \emph{ST-operator}.
    \item RNN+FL, MTrajRec+FL, and RNTrajRec+FL can capture the temporal dependencies, RNTrajRec+FL performs the best. This is because RNTrajRec is capable of learning more effective trajectory embeddings benefiting from the graph model, which incorporates the embedding of each road segment.
\end{itemize}
The above observations indicate that LightTR is more effective than the existing FL-based methods.

\begin{table}
    \footnotesize
    \centering
    \setlength\tabcolsep{7pt}
    \renewcommand\arraystretch{1.0}
    \caption{Effect of the Number of Clients on Both Datasets (keep ratio=12.5\%)}
    \vspace{-0.3cm}
    \begin{tabular}{c|c|c|c|c|c}
    \hline
    \multirow{2}{*}{\textbf{Datasets}} & \multirow{2}{*}{\textbf{Metrics}} & \multicolumn{4}{c}{\textbf{Number of Clients}}\\
    \cline{3-6} & & 5 & 10 & 15 & 20\\
    \hline
    \multirow{4}{*}{\textbf{Geolife}}
    &\textbf{Recall}  &0.629 & 0.629 &\textbf{0.741} &0.724\\
    &\textbf{Precision} &0.613 &0.692 &0.731 &\textbf{0.748}\\
    &\textbf{MAE} &0.444 &0.451 &0.377 &\textbf{0.335}\\
    &\textbf{RMSE} &0.524 &0.536 & 0.478 &\textbf{0.432}\\
    \hline
    \hline
    \multirow{4}{*}{\textbf{Tdrive}}
    &\textbf{Recall} &0.622 &0.631 &0.657 &\textbf{0.666}\\
    &\textbf{Precision} &0.630 &0.690 &0.700 &\textbf{0.707}\\
    &\textbf{MAE} &0.452 &449 &\textbf{0.355} &0.366\\
    &\textbf{RMSE} &0.554 &561 &0.461 &\textbf{0.452}\\
    \hline
    \end{tabular}
    \vspace{-0.5cm}
    \label{Clients}
\end{table}

\begin{table*}
    \footnotesize
    \centering
    \caption{Centralized Method vs. LightTR}
    \vspace{-0.3cm}
    \renewcommand\arraystretch{1.0}
    \setlength\tabcolsep{5pt}
    \begin{tabular}{c|c|c|c|c|c|c|c|c|c|c|c|c|c}
    \hline
    \multirow{2}{*}{\textbf{Datasets}} & \multirow{2}{*}{\textbf{Baseline}} & 
    \multicolumn{4}{c|}{\textbf{6.25\%}} & \multicolumn{4}{c|}{\textbf{12.5\%}} &
    \multicolumn{4}{c}{\textbf{25\%}}\\
        \cline{3-14} & & Recall & Precision & MAE & RMSE & Recall & Precision & MAE & RMSE & Recall & Precision & MAE & RMSE \\
        \hline
        \multirow{2}{*}{\textbf{Geolife}}
        &\textbf{MTrajRec} & {0.697} & {0.781}  & {0.315}  & {0.459} & 0.723 &0.739 &0.337 & 0.502 & 0.732 & 0.730 & 0.376& 0.461\\
        &\textbf{LightTR} & {0.674}& {0.679} & {0.375} & {0.464} &{0.724} &{0.748} &{0.335} &{0.432} &{0.758} &{0.739} &{0.323} &{0.435}\\
        \hline
        \hline
        \multirow{2}{*}{\textbf{Tdrive}}
        &\textbf{MTrajRec} &{0.630} & {0.678}& 0.421 &0.593 &0.631 &0.676 &0.411 &0.619 &0.693 &0.711 &0.368 &0.481\\   
        &\textbf{LightTR} &{0.624} &{0.674} &{0.395} &{0.513} &{0.667} &{0.707} &{0.366} &{0.452} &{0.708} &{0.732} &{0.332} &{0.446}\\
        \hline
        \end{tabular}
        \vspace{-0.4cm}
        \label{centralizedvs}
\end{table*}

\subsubsection{Effect of the Number of Clients.}
To study the effect of the number of clients involved in model training of the proposed LightTR, we conduct experiments with 5, 10, 15, and 20 clients. The results are shown in Table~\ref{Clients}. We observe that the model performance gets better as the increase of number of clients, due to more clients bringing more training data. For example, the Recall on Geolife increases from 0.629 to 0.724, and the Precision on Tdrive increases from 0.630 to 0.707. One can also see that the 20-client-based LightTR performs worse than the 15-client-based LightTR in terms of Recall. Generally, the results demonstrate that more clients participating in training are more like to lead to better performance because of more useful knowledge is learned from more training data during the training.

\begin{figure}[h]
    \centering
    \vspace{-0.4cm}
    \subfigure[Running time]{
    \label{Fig.sub.1}
    \includegraphics[scale=0.25]{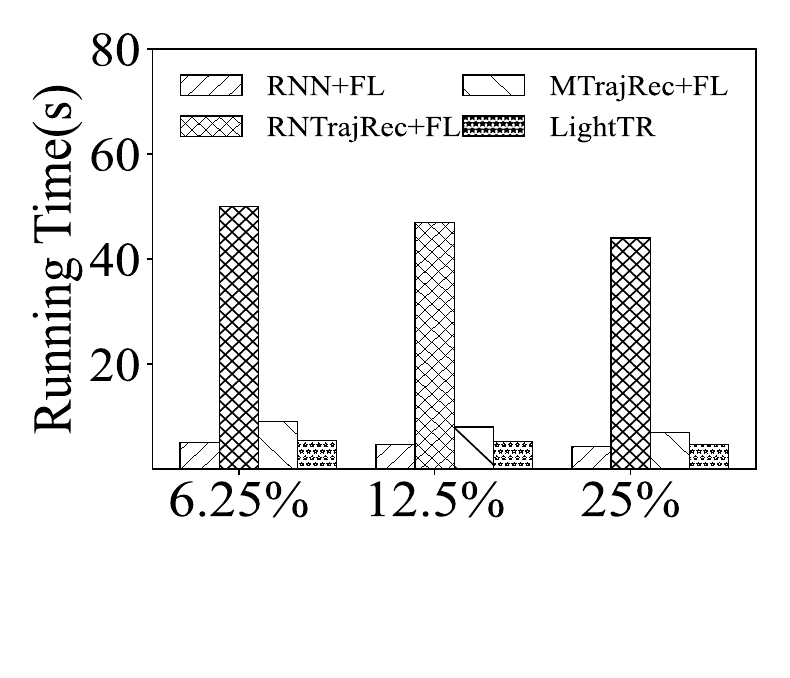}}
    \subfigure[FLOPs and Parameters]{
    \label{Fig.sub.2}
    \includegraphics[scale=0.24]{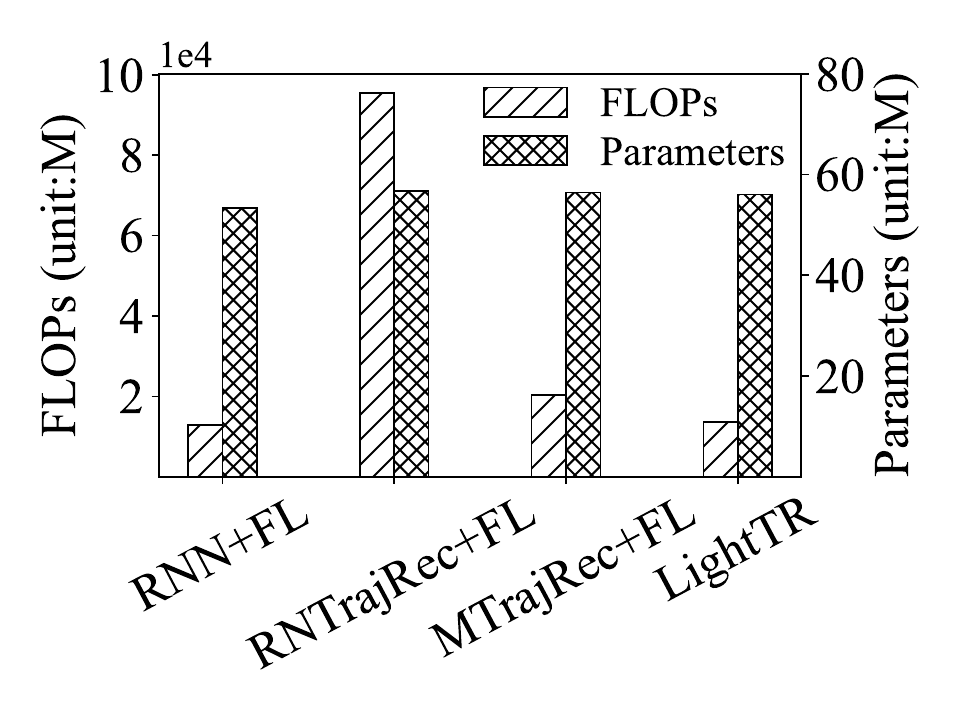}}
    \vspace{-0.2cm}
    \caption{Running Efficiency on Geolife}
    \vspace{-0.2cm}
    \label{Efficiency}
\end{figure}

\subsubsection{Running Efficiency.}
Considering lightness and scalability, we study the running time (of each epoch), floating-point operations (FLOPs), and the number of parameters for LightTR with three RNN-based baselines (i.e., RNN+FL, MTrajRec+FL, and RNTrajRec+FL) on Geolife. Figure~\ref{Efficiency}(a) shows the results of running time, while FLOPs and parameters consumptions are shown in Figure~\ref{Efficiency}(b). Generally, LightTR has significantly less running time and fewer FLOPs due to the lightness of the proposed lightweight \emph{ST-operator}. We also observe an exception in that RNN+FL achieves slightly less running time. However, RNN+FL is much less accurate. In addition, LightTR achieves faster overall convergence, which offers evidence that the proposed meta-knowledge enhanced local-global training module can accelerate model training. Specifically, LightTR converges after around 100 epochs, while the second fastest baseline (except RNN+FL) MtrajRec+FL converges after around 160 epochs, on both datasets. It is noteworthy that we include the training convergence curve at \url{https://github.com/uestc-liuzq/LightTR} to save space.

It is clear that LightTR uses fewer resources than the two most accurate competing models MTrajeRec+FL and RNTrajRec, while maintaining comparable accuracy. More specifically, LightTR reduces the running time by 90\% and 32.9\% compared with RNTrajRec+FL and MTrajRec, respectively. Additionally, LightTR obtains FLOPs reduction by 86.7\% compared to RNTrajRec as the high calculation complexities of the attention mechanism in RNTrajRec. Note that the communication cost in FL is positively correlated with the number of model parameters and FLOPs~\cite{chen2018federated, feng2020pmf}, where FLOPs are termed as system overhead. Thus, the fewer FLOPs and parameters consumed by LightTR bring the benefits of fewer communication costs and less system overhead in federated trajectory recovery. Overall, the results in Figure~\ref{Efficiency} indicate the feasibility and scalability of LightTR for model deployment in real decentralized trajectory recovery scenarios.

\subsubsection{Effect of Client Fractions.}  
\begin{figure}[h]
    \centering
    \vspace{-0.4cm}
    \subfigure[Geolife]{
    \label{Fig.sub.1}
    \includegraphics[scale=0.25]{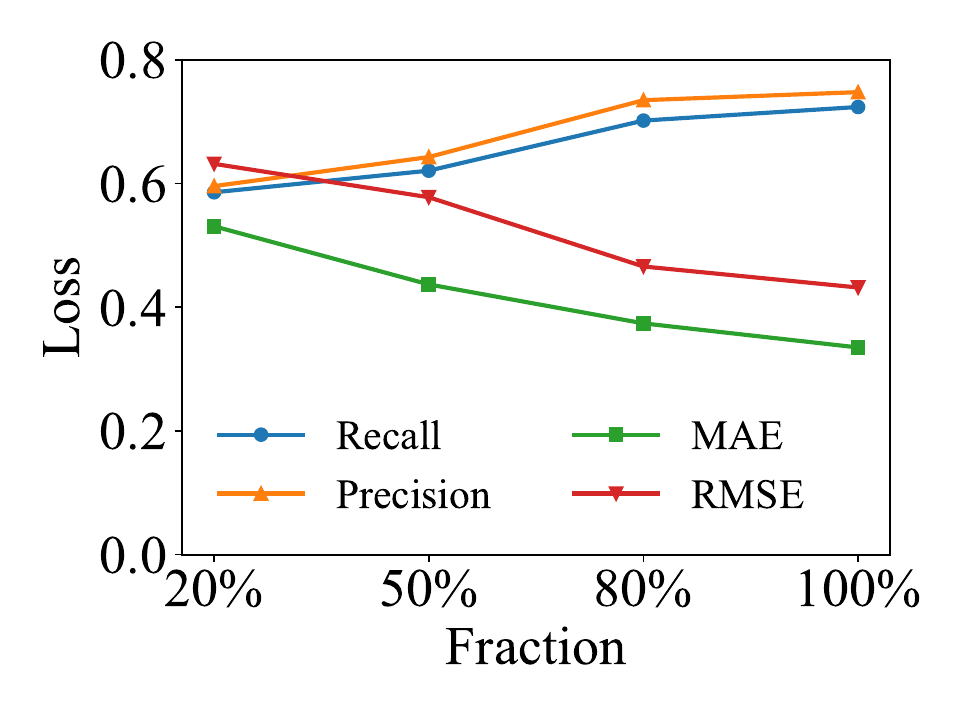}}
    \subfigure[Tdrive]{
    \label{Fig.sub.1}
    \includegraphics[scale=0.25]{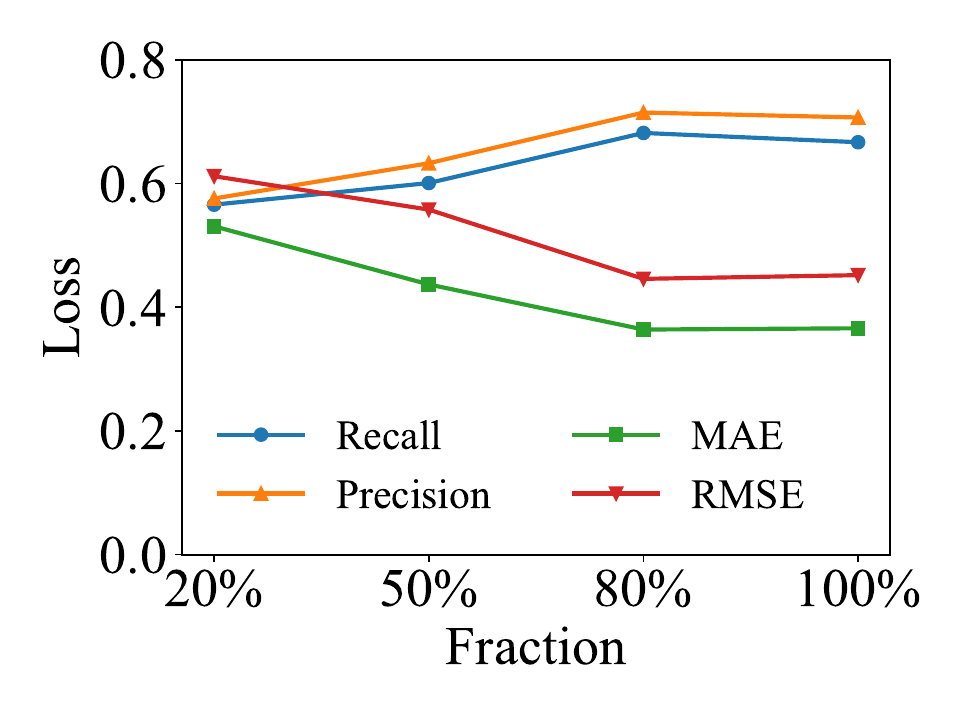}}
    \vspace{-0.2cm}
    \caption{Effect of Client Fractions on Both Datasets (keep ratio=12.5\%)}
    \vspace{-0.3cm}
    \label{Fractions}
\end{figure}

When we aggregate parameters in LightTR, a certain fraction of clients is sampled to perform model training in each client. To study the effect of client fraction $F$ on the model performance, we conduct experiments by sampling four different fractions of clients: 20\%, 50\%, 80\%, and 100\%, as shown in Figure~\ref{Fractions}. Generally, the model performance exhibits an increasing trend with the increase in client fractions in terms of all evaluation metrics. LightTR achieved a most improvement of 35.8\% in recall and 37.2\% in precision, respectively, while getting MAE and RMSE most reduction by up to 59.4\% and 50.7\%. It indicates that the bigger fraction of clients with more training data is, the better model performs since more useful knowledge can be learned.

\subsubsection{Centralized Method vs. LightTR}
To compare the proposed LightTR with the centralized method in terms of trajectory recovery, we select MTrajRec~\cite{r2}, a recent stat-of-the-art deep learning framework for centralized trajectory recovery, which integrates attention mechanism with Seq2Seq to learn effective trajectory embeddings. Although the centralized model reduces privacy severely, which makes the learned model vulnerable, LightTR performs better than MTrajRec in most cases, as shown in Table~\ref{centralizedvs}. This is because LightTR learns local meta-knowledge by means of knowledge distillation, which is helpful for better trajectory embedding learning. In addition, we observe that LightTR gets more performance improvement compared with MTrajRec on Tdrive. The reason is that data contained in Tdrive is sparse, and the proposed meta-knowledge local-global training module can handle the sparsity and learn more effective trajectory embeddings.

\subsubsection{Ablation Study.}
\begin{figure}[t]
\centering  
\subfigure[Recall]{
\label{Fig.sub.1}
\includegraphics[scale=0.25]{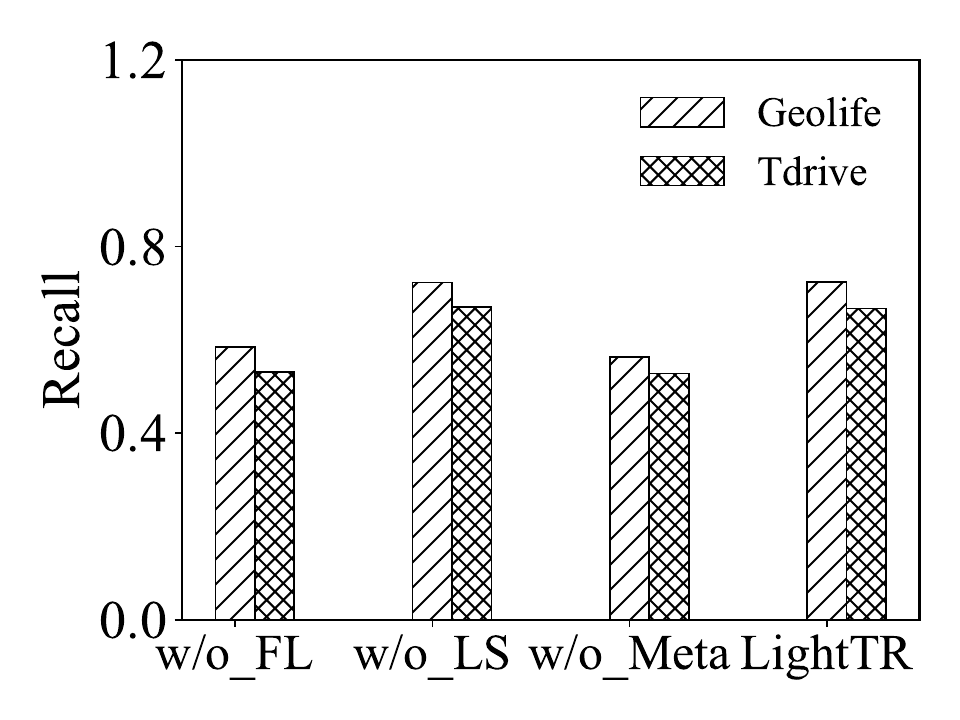}}
\subfigure[Precision]{
\label{Fig.sub.2}
\includegraphics[scale=0.25]{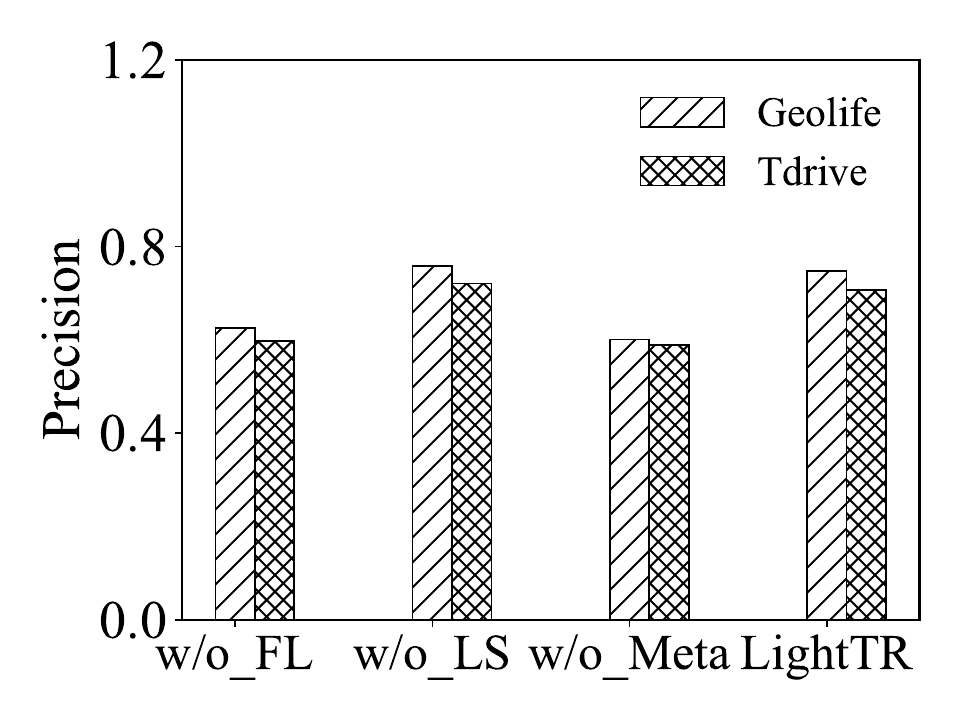}}
\subfigure[MAE]{
\label{Fig.sub.3}
\includegraphics[scale=0.25]{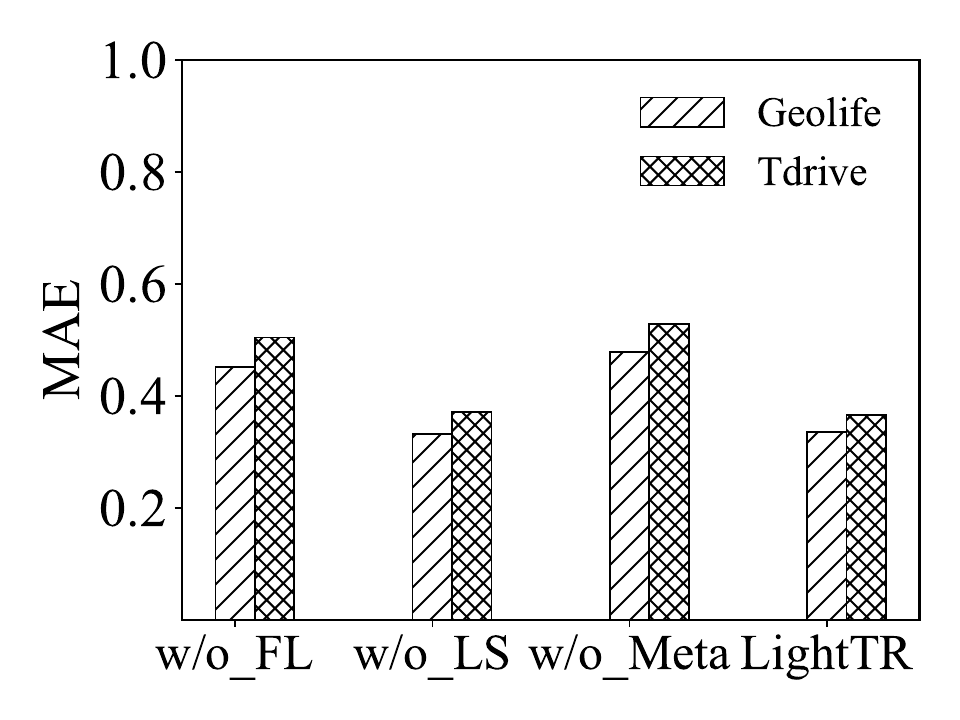}}
\subfigure[RMSE]{
\label{Fig.sub.4}
\includegraphics[scale=0.25]{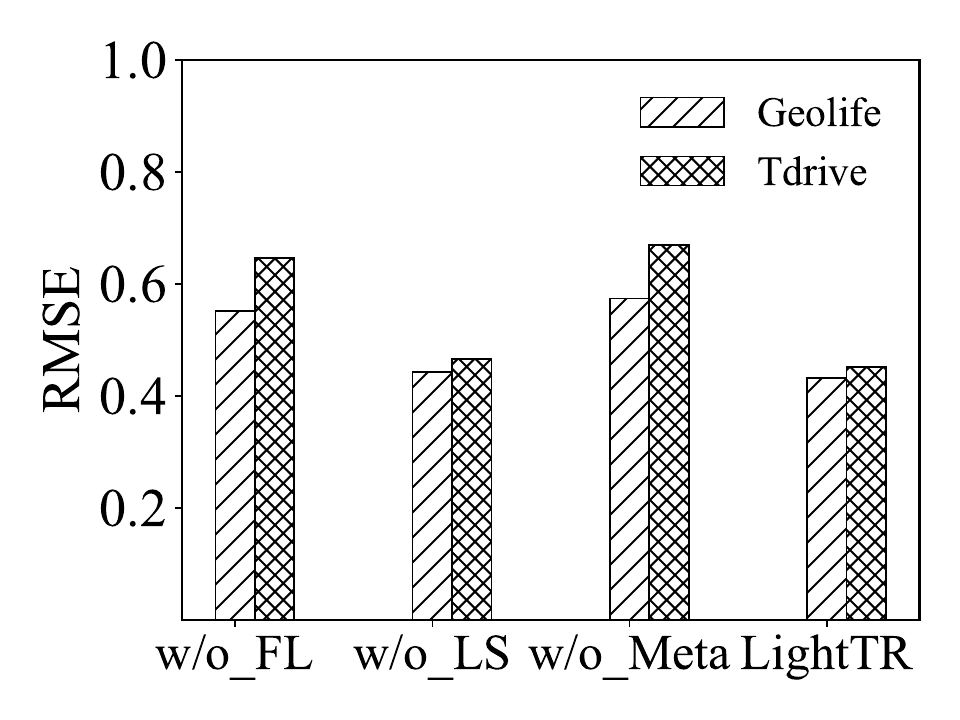}}
\vspace{-0.2cm}
\caption{Performance of LightTR and Its Variants on Both Datasets (keep ratio = 12.5\%).}
\vspace{-0.15cm}
\label{components}
\end{figure}

To gain insight into the effects of key aspects of LightTR, We evaluate three LightTR variants.
\begin{itemize}
    \item \textbf{w/o horizontal FL (w/o\_FL).} 
    w/o\_FL removes the central server of LightTR, where each client trains their model locally and all clients exchange their local model to each other.
    \item \textbf{w/o Lightweight \emph{ST-operator} (w/o\_LS).} w/o\_LS replace the Lightweight \emph{ST-operator} with MTrajRec to learn local trajectory embeddings for each client.
    \item \textbf{w/o meta-knowledge enhanced local-global training module (w/o\_Meta).} w/o\_Meta replace the meta-knowledge enhanced local-global training module with FedAvg~\cite{r4}.
\end{itemize}
To assess whether the components in LightTR all contribute to the performance of trajectory recovery, we compare LightTR with its variants. The results are shown in Figure~\ref{components}. The observations are as follows.
\begin{itemize}
    \item 
    Regardless of the datasets, LightTR always performs better than its counterparts without horizontal FL, lightweight \emph{ST-operator}, and meta-knowledge enhanced local-global training module. It shows that these components are all useful for trajectory recovery.
    \item 
    LightTR performs slightly better than w/o\_LS. This shows the powerful feature learning capability of the proposed lightweight \emph{ST-operator}, even compared with MTrajRec, a recent state-of-the-art trajectory recovery model. However, the complex model always has high space and time complexities, which incurs poor scalabilities and more computation resource requirements.
    \item 
    w/o\_Meta performs the worst among all variants, which shows the importance of the meta-knowledge enhanced local-global training module. It also offers evidence that the meta-knowledge enhanced local-global training module is capable of alleviating the data heterogeneities (Non-IIDness) across clients and enhancing effective trajectory learning.
\end{itemize} 

\subsubsection{Parameter Sensitivities.} 
\begin{figure}[h]
\centering  
\vspace{-0.5cm}
\subfigure[Effect of $\lambda$]{
\label{lambda}
\includegraphics[scale=0.25]{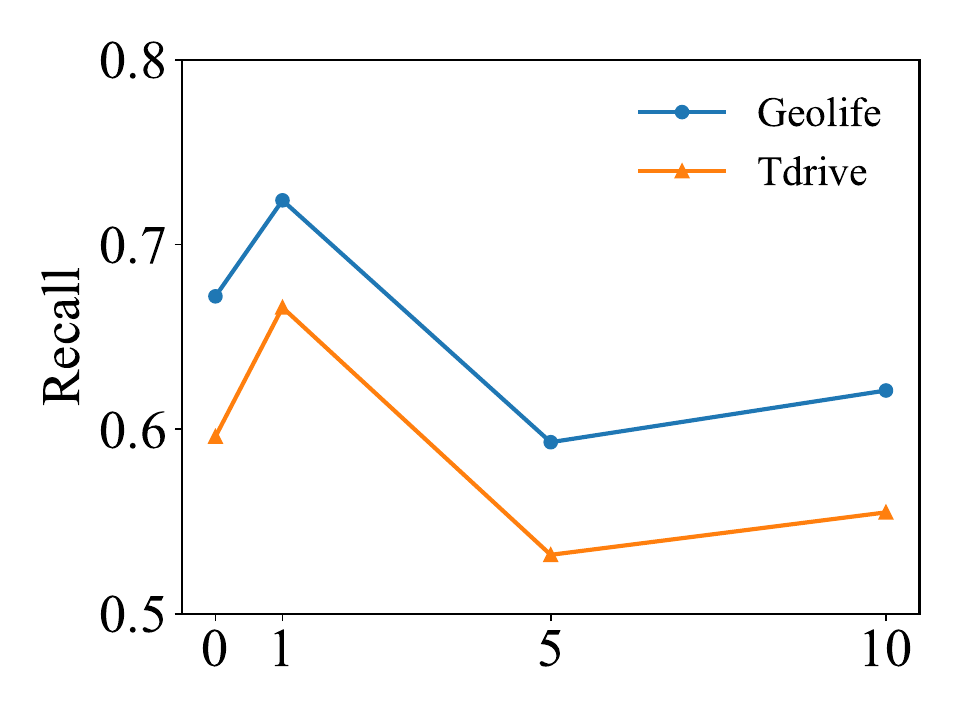}
\includegraphics[scale=0.25]{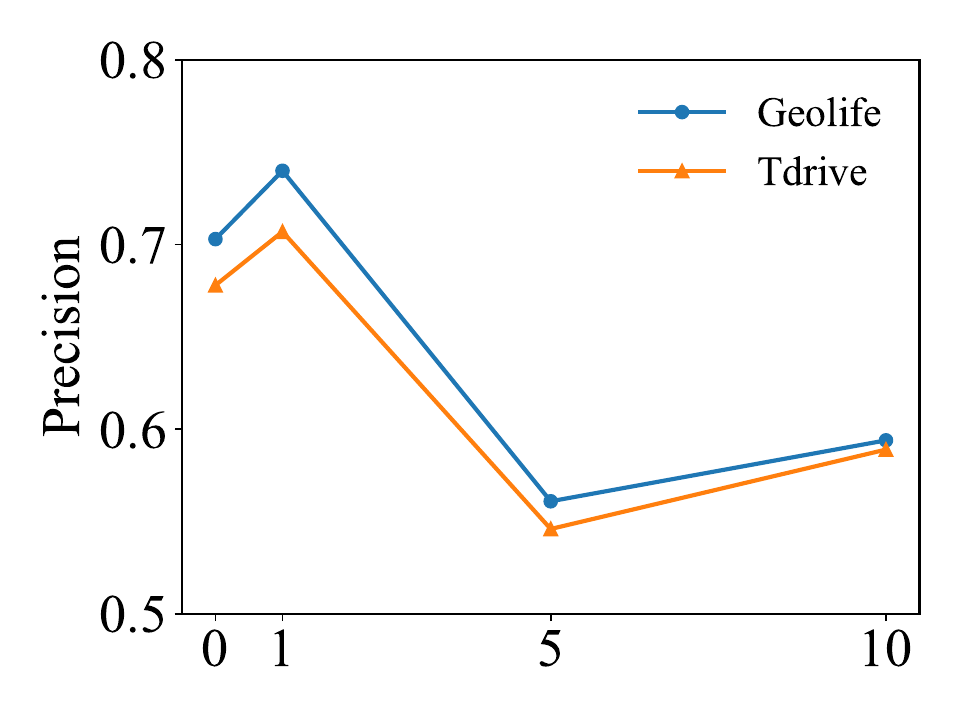}}
\vspace{-0.2cm}
\subfigure[Effect of $l_t$]{
\label{threshold}
\includegraphics[scale=0.25]{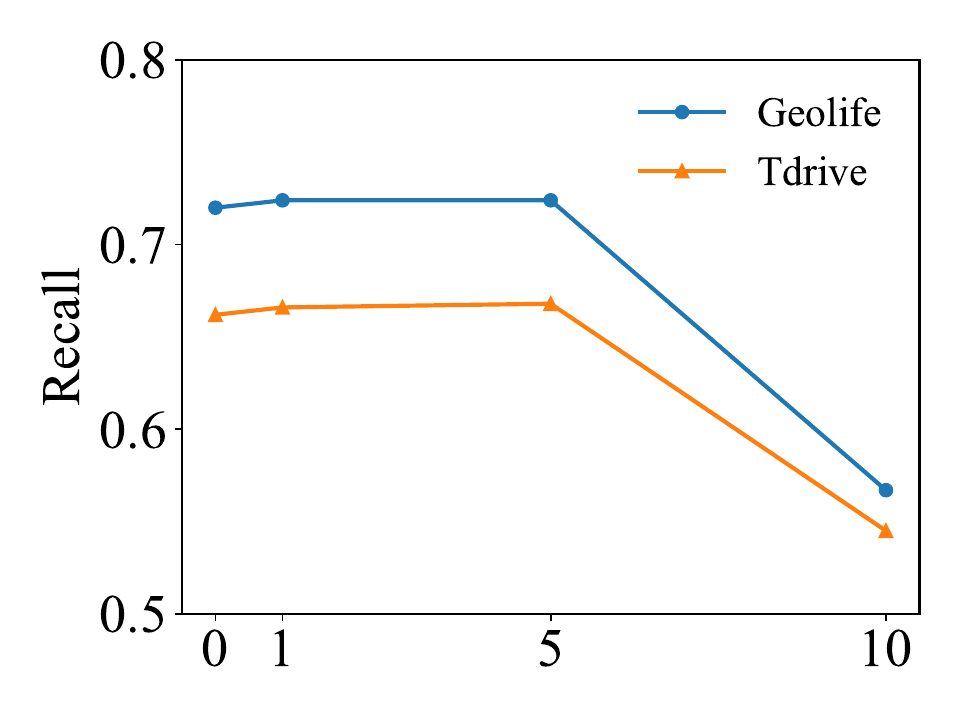}
\includegraphics[scale=0.25]{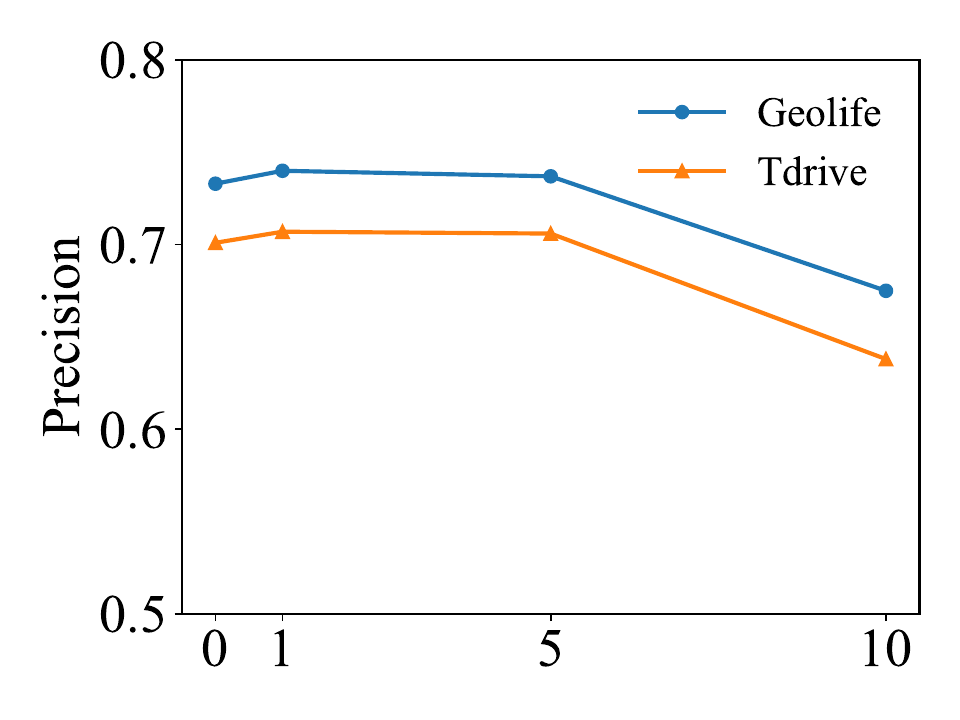}}
\caption{Parameters Sensitivity Analysis on both datasets (keep ratio = 12.5\%)}
\vspace{-0.8cm}
\label{Parameters}
\end{figure}

We next study how sensitive the model is on the hyper-parameter $\lambda$ and the threshold $l_t$, as shown in Figure~\ref{Parameters}. $\lambda$ is set to different values, i.e., 0.1, 1, 5, 10. Figure~\ref{Parameters}(a) illustrates the effect of $\lambda$, showing setting $\lambda$ to 5 is the reasonable setting in this experiment. It indicates that the learned meta-knowledge plays a more important role in trajectory modeling, which shows the necessity of the proposed meta-knowledge enhanced local-global training module. 
Figure~\ref{Parameters}(b) shows the effect of $l_t$, which is the threshold for knowledge accumulation, and we set it from 0 to 0.6. Here, we observe all curves slightly increase and then decrease. LightTR achieves the best result when setting $l_t$ to 0.4. It indicates that excessive guidance from the meta-learner will make the local trajectory embedding model confusing, and thus degrade the trajectory recovery performance.

\begin{figure*}[ht]
\centering  
\subfigure[Ground Truth]{
\centering
\label{ground}
\includegraphics[scale=0.26]{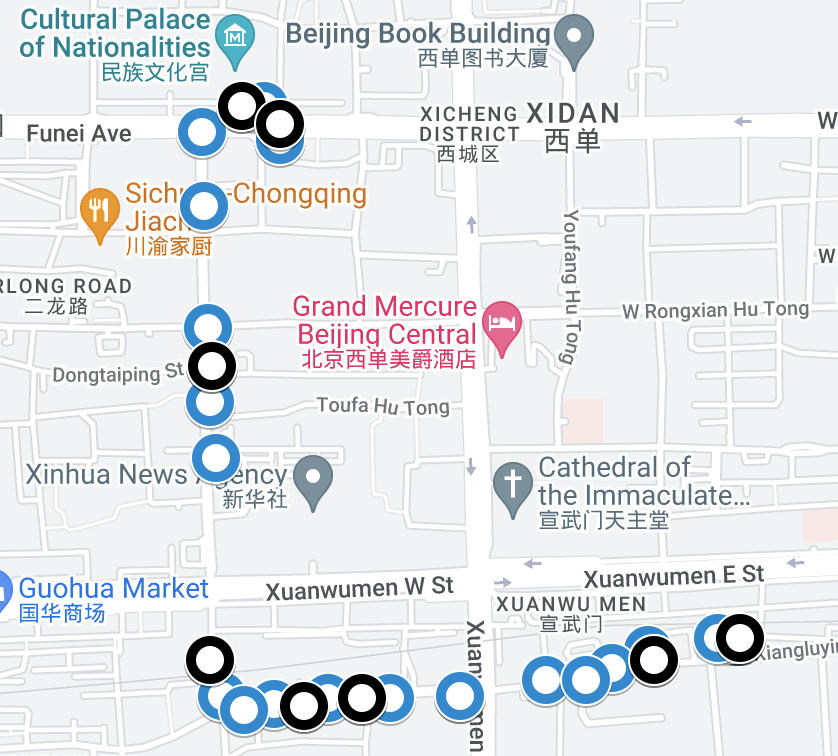}}
\subfigure[LightTR]{
\label{prediction}
\includegraphics[scale=0.26]{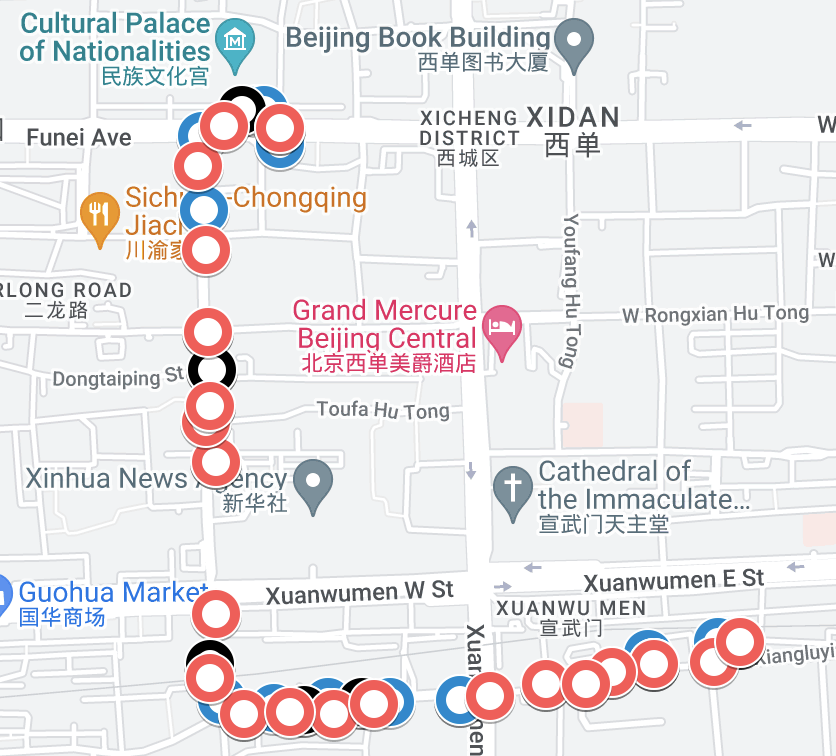}
}
\subfigure[RNN+FL]{
\label{RNN}
\includegraphics[scale=0.26]{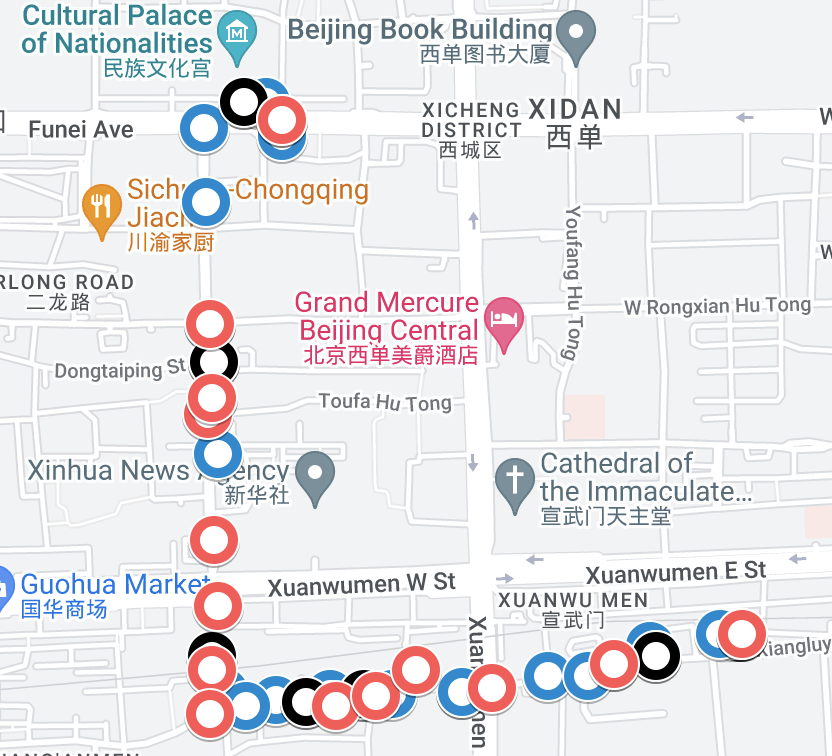}}
\subfigure[RNTrajRec+FL]{
\label{RNTrajRec}
\includegraphics[scale=0.26]{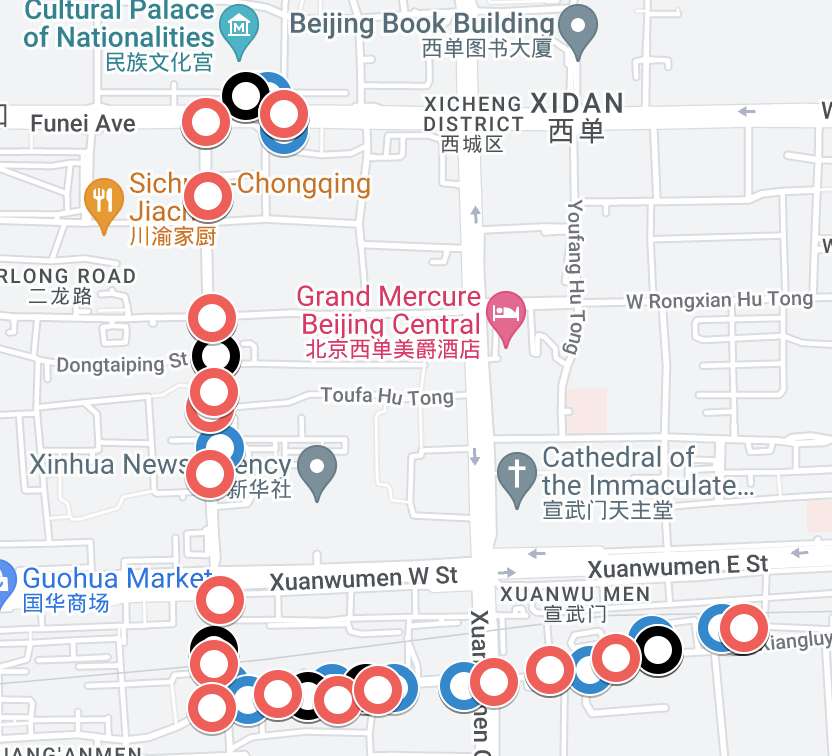}}
\vspace{-0.2cm}
\caption{Ground truth vs. prediction on Tdrive. Black points represent the low-sampling-rate trajectory. Red points denote the prediction of recovered points, and blue points are the ground truth of high-sampling-rate trajectory (keep ratio = 12.5\%).}
\label{CaseStudy}
\vspace{-0.5cm}
\end{figure*}

\subsubsection{Case Study}
To further intuitively illustrate how accurately LightTR can recover low-sampling-rate trajectories (black points), we visualize the predicted points (in red) and the ground truth (in blue) at a keep ratio 12.5\% compared with two baseline models (RNN+FL, RNTrajRec+FL), as depicted in Figure~\ref{CaseStudy}. Figure~\ref{CaseStudy}(b) shows the recovered points exhibit a remarkable level of alignment with the ground truth, which is largely due to the usage of the proposed meta-knowledge enhanced local-global training module by means of knowledge distillation. It is clear that LightTR can accurately trace the right route and the recovered points are more reliable than the other two baselines. Figure~\ref{CaseStudy}(c) shows the recovery results of RNN+FL. Although it roughly predicts the right trajectory route, most points are not correctly recovered especially in the bottom area. This is because simply stacked RNNs cannot learn effective trajectory embeddings, and thus suffer from poor accuracy when recovering a large number of missing points. Figure~\ref{CaseStudy}(d) shows the recovery results of RNTrajRec+FL. It finds the right path as the left two methods but the recovered points are more accurate than RNN+FL since RNTrajRec utilizes a powerful network (i.e., graph model and transformers) to learn more effective features of trajectories than RNN+FL.
\section{Related Work}
\label{relatedwork}
We briefly review prior studies on trajectory recovery and federated learning.
\subsection{Trajectory Recovery}

Trajectory recovery~\cite{r14, r2, r1, zhang2023simidtr, chen2023teri} attracts increasing interest due to the increasing availability of trajectory data and rich downstream applications in intelligent transportation systems. 

Traditionally, trajectory recovery methods are mostly based on statistical models~\cite{r14, r22}. InferTra~\cite{r14} learns a Network Mobility Model (NMM) from historical trajectories, which employ Gibbs sampling for trajectory recovery. 
However, the statistical models cannot capture complex temporal (and spatial) correlations of trajectory data due to their limited learning capacity.

With the advance of deep learning techniques, recent deep learning based methods~\cite{r1, r13, r2, r3} address trajectory recovery by leveraging their powerful learning capabilities. 
For example, MTrajRec~\cite{r2} addresses constrained trajectory restoration through Seq2Seq based multi-task learning. However, these methods are trained with centralized data, while disregard decentralized settings and do not take computational efficiency into account.


\subsection{Federated Learning}

Federated Learning (FL) is a machine learning approach where many clients (often called edge devices) collaboratively train a model using decentralized data under the orchestration of a central server~\cite{r7, r4}. Generally, FL can be divided into three categories: horizontal FL~\cite{r4}, vertical FL~\cite{r27}, and federated transfer learning~\cite{r29}. Moreover, data heterogeneity across clients may degrade the performance of FL. To solve this problem, personalized FL~\cite{r10, r11} is widely studied recently. For example, FedDC~\cite{r10} introduces local drift decoupling and correction to address Non-IID data across clients. 
However, most of the above methods are designed for computer vision, and cannot be applied to trajectory data due to the unique spatio-temporal patterns.

Recently, FL is applied in spatio-temporal data (e.g., traffic data) to ensure data privacy~\cite{r6,r12,r41}. For instance, CNFGNN~\cite{r6} is proposed to perform federated traffic prediction, which extracts high-dimensional spatial and temporal features utilizing GNN and GRU. Additionally, FRA~\cite{r12} explores spatial data aggregation queries in federated scenarios. Meanwhile, FL has also been applied to the trajectory data. 
FedLoc~\cite{r44} collaborates to build accurate positioning services without sacrificing user privacy, particularly sensitive information related to its geographic trajectory.

However, these methods are computational heavily. There still lacks a well-customized FL model for trajectory recovery, considering the effectiveness and efficiency at the same time.

\section{CONCLUSION}
\label{conclusion}


We present LightTR, a new federated framework for lightweight trajectory recovery that aims at decentralized trajectory learning and privacy protection. To reduce the computational cost, we design a lightweight local trajectory embedding module with a lightweight \emph{ST-operator} for each client, which is also capable of effective feature extraction. A novel meta-knowledge enhanced local-global training module is proposed to offer improved communication efficiency between the server and clients, which alleviates the data heterogeneity across clients. Comprehensive experiments on two real datasets offer evidence that LightTR achieves state-of-the-art accuracy but consumes fewer computational resources in trajectory recovery.


\section{Acknowledgment}
This work is partially supported by NSFC (No. 61972069, 61836007 and 61832017),  Shenzhen Municipal Science and Technology R\&D Funding Basic Research Program (JCYJ20210324133607021), and Municipal Government of Quzhou under Grant (No. 2022D037, 2023D044), and Key Laboratory of Data Intelligence and Cognitive Computing, Longhua District, Shenzhen. In addition, Chenxi Liu participated in this work when she was at Hunan University.


\bibliographystyle{IEEEtran}
\bibliography{ref}

\begin{thebibliography}{10}
\providecommand{\url}[1]{#1}
\csname url@samestyle\endcsname
\providecommand{\newblock}{\relax}
\providecommand{\bibinfo}[2]{#2}
\providecommand{\BIBentrySTDinterwordspacing}{\spaceskip=0pt\relax}
\providecommand{\BIBentryALTinterwordstretchfactor}{4}
\providecommand{\BIBentryALTinterwordspacing}{\spaceskip=\fontdimen2\font plus
\BIBentryALTinterwordstretchfactor\fontdimen3\font minus
  \fontdimen4\font\relax}
\providecommand{\BIBforeignlanguage}[2]{{%
\expandafter\ifx\csname l@#1\endcsname\relax
\typeout{** WARNING: IEEEtran.bst: No hyphenation pattern has been}%
\typeout{** loaded for the language `#1'. Using the pattern for}%
\typeout{** the default language instead.}%
\else
\language=\csname l@#1\endcsname
\fi
#2}}
\providecommand{\BIBdecl}{\relax}
\BIBdecl

\bibitem{luo2013finding}
W.~Luo, H.~Tan, L.~Chen, and L.~M. Ni, ``Finding time period-based most
  frequent path in big trajectory data,'' in \emph{SIGMOD}, 2013, pp. 713--724.

\bibitem{DBLP:conf/pkdd/ZhangWWZMZ22}
S.~Zhang, S.~Wang, X.~Wang, S.~Zhang, H.~Miao, and J.~Zhu, ``Multi-task
  adversarial learning for semi-supervised trajectory-user linking,'' in
  \emph{{ECML}-{PKDD}}, vol. 13716, 2022, pp. 418--434.

\bibitem{zhao2018rest}
Y.~Zhao, S.~Shang, Y.~Wang, B.~Zheng, Q.~V.~H. Nguyen, and K.~Zheng, ``Rest: A
  reference-based framework for spatio-temporal trajectory compression,'' in
  \emph{SIGKDD}, 2018, pp. 2797--2806.

\bibitem{DBLP:conf/kdd/LiuSZZZ21}
S.~Liu, H.~Su, Y.~Zhao, K.~Zeng, and K.~Zheng, ``Lane change scheduling for
  autonomous vehicle: {A} prediction-and-search framework,'' in \emph{SIGKDD},
  2021, pp. 3343--3353.

\bibitem{chen2022modeling}
M.~Chen, Y.~Zhao, Y.~Liu, X.~Yu, and K.~Zheng, ``Modeling spatial trajectories
  with attribute representation learning,'' \emph{TKDE}, vol.~34, no.~4, pp.
  1902--1914, 2022.

\bibitem{zheng2019reference}
K.~Zheng, Y.~Zhao, D.~Lian, B.~Zheng, G.~Liu, and X.~Zhou, ``Reference-based
  framework for spatio-temporal trajectory compression and query processing,''
  \emph{TKDE}, vol.~32, no.~11, pp. 2227--2240, 2019.

\bibitem{deng2023s2tul}
L.~Deng, H.~Sun, Y.~Zhao, S.~Liu, and K.~Zheng, ``S2tul: A semi-supervised
  framework for trajectory-user linking,'' in \emph{WSDM}, 2023, pp. 375--383.

\bibitem{xiao2021vehicle}
J.~Xiao, Z.~Xiao, D.~Wang, V.~Havyarimana, C.~Liu, C.~Zou, and D.~Wu, ``Vehicle
  trajectory interpolation based on ensemble transfer regression,''
  \emph{TITS}, vol.~23, no.~7, pp. 7680--7691, 2021.

\bibitem{r6}
C.~Meng, S.~Rambhatla, and Y.~Liu, ``Cross-node federated graph neural network
  for spatio-temporal data modeling,'' in \emph{SIGKDD}, 2021, pp. 1202--1211.

\bibitem{r23}
Y.~Tong, D.~Shi, Y.~Xu, W.~Lv, Z.~Qin, and X.~Tang, ``Combinatorial
  optimization meets reinforcement learning: Effective taxi order dispatching
  at large-scale,'' \emph{TKDE}, pp. 1--1, 2021.

\bibitem{zheng2012reducing}
K.~Zheng, Y.~Zheng, X.~Xie, and X.~Zhou, ``Reducing uncertainty of
  low-sampling-rate trajectories,'' in \emph{ICDE}.\hskip 1em plus 0.5em minus
  0.4em\relax IEEE, 2012, pp. 1144--1155.

\bibitem{r14}
P.~Banerjee, S.~Ranu, and S.~Raghavan, ``Inferring uncertain trajectories from
  partial observations,'' in \emph{ICDM}, 2014, pp. 30--39.

\bibitem{r15}
J.~Yuan, Y.~Zheng, C.~Zhang, X.~Xie, and G.-Z. Sun, ``An interactive-voting
  based map matching algorithm,'' in \emph{MDM}, 2010, pp. 43--52.

\bibitem{r22}
J.~A. Alvarez-Garcia, J.~A. Ortega, L.~Gonzalez-Abril, and F.~Velasco, ``Trip
  destination prediction based on past gps log using a hidden markov model,''
  \emph{Expert Systems with Applications}, vol.~37, no.~12, pp. 8166--8171,
  2010.

\bibitem{r1}
H.~Sun, C.~Yang, L.~Deng, F.~Zhou, F.~Huang, and K.~Zheng, ``Periodicmove:
  Shift-aware human mobility recovery with graph neural network,'' in
  \emph{CIKM'21}, 2021, pp. 1734--1743.

\bibitem{r2}
H.~Ren, S.~Ruan, Y.~Li, J.~Bao, C.~Meng, R.~Li, and Y.~Zheng, ``Mtrajrec:
  Map-constrained trajectory recovery via seq2seq multi-task learning,'' in
  \emph{SIGKDD}, 2021, pp. 1410--1419.

\bibitem{r3}
T.~Xia, Y.~Qi, J.~Feng, F.~Xu, F.~Sun, D.~Guo, and Y.~Li, ``Attnmove: History
  enhanced trajectory recovery via attentional network,'' in \emph{AAAI},
  vol.~35, no.~5, 2021, pp. 4494--4502.

\bibitem{yang2023long}
C.~Yang and Z.~Pei, ``Long-short term spatio-temporal aggregation for
  trajectory prediction,'' \emph{TITS}, vol.~24, no.~4, pp. 4114--4126, 2023.

\bibitem{r13}
J.~Wang, N.~Wu, X.~Lu, W.~X. Zhao, and K.~Feng, ``Deep trajectory recovery with
  fine-grained calibration using kalman filter,'' \emph{TKDE}, vol.~33, no.~3,
  pp. 921--934, 2019.

\bibitem{r7}
K.~Wei, J.~Li, M.~Ding, C.~Ma, H.~H. Yang, F.~Farokhi, S.~Jin, T.~Q.~S. Quek,
  and H.~Vincent~Poor, ``Federated learning with differential privacy:
  Algorithms and performance analysis,'' \emph{IEEE T INF FOREN SEC}, vol.~15,
  pp. 3454--3469, 2020.

\bibitem{r4}
B.~McMahan, E.~Moore, D.~Ramage, S.~Hampson, and B.~A.~y. Arcas,
  ``{Communication-Efficient Learning of Deep Networks from Decentralized
  Data},'' in \emph{AISTATS}, 2017, pp. 1273--1282.

\bibitem{lai2023lightcts}
Z.~Lai, D.~Zhang, H.~Li, C.~S. Jensen, H.~Lu, and Y.~Zhao, ``Lightcts: A
  lightweight framework for correlated time series forecasting,''
  \emph{SIGMOD}, vol.~1, no.~2, pp. 1--26, 2023.

\bibitem{brakatsoulas2005map}
S.~Brakatsoulas, D.~Pfoser, R.~Salas, and C.~Wenk, ``On map-matching vehicle
  tracking data,'' in \emph{PVLDB}, 2005, pp. 853--864.

\bibitem{rappos2018force}
E.~Rappos, S.~Robert, and P.~Cudr{\'e}-Mauroux, ``A force-directed approach for
  offline gps trajectory map matching,'' in \emph{SIGSPATIAL}, 2018, pp.
  319--328.

\bibitem{chambers2020map}
E.~Chambers, B.~T. Fasy, Y.~Wang, and C.~Wenk, ``Map-matching using shortest
  paths,'' \emph{TSAS}, vol.~6, no.~1, pp. 1--17, 2020.

\bibitem{r17}
S.~Zheng, Y.~Yue, and P.~Lucey, ``Generating long-term trajectories using deep
  hierarchical networks,'' in \emph{NIPS}, 2016, p. 1551–1559.

\bibitem{chen2011discovering}
Z.~Chen, H.~T. Shen, and X.~Zhou, ``Discovering popular routes from
  trajectories,'' in \emph{ICDE}, 2011, pp. 900--911.

\bibitem{su2013calibrating}
H.~Su, K.~Zheng, H.~Wang, J.~Huang, and X.~Zhou, ``Calibrating trajectory data
  for similarity-based analysis,'' in \emph{SIGMOD}, 2013, pp. 833--844.

\bibitem{su2015calibrating}
H.~Su, K.~Zheng, J.~Huang, H.~Wang, and X.~Zhou, ``Calibrating trajectory data
  for spatio-temporal similarity analysis,'' \emph{VLDBJ}, vol.~24, pp.
  93--116, 2015.

\bibitem{lun2023resisting}
Y.~Lun, H.~Miao, J.~Shen, R.~Wang, X.~Wang, and S.~Wang, ``Resisting tul
  attack: balancing data privacy and utility on trajectory via collaborative
  adversarial learning,'' \emph{GeoInformatica}, pp. 1--21, 2023.

\bibitem{fang2023heterogeneous}
J.~Fang, C.~Zhu, P.~Zhang, H.~Yu, and J.~Xue, ``Heterogeneous trajectory
  forecasting via risk and scene graph learning,'' \emph{TITS}, 2023.

\bibitem{chen2020multi}
W.~Chen, L.~Chen, Y.~Xie, W.~Cao, Y.~Gao, and X.~Feng, ``Multi-range attentive
  bicomponent graph convolutional network for traffic forecasting,'' in
  \emph{AAAI}, vol.~34, no.~04, 2020, pp. 3529--3536.

\bibitem{wu2021autocts}
X.~Wu, D.~Zhang, C.~Guo, C.~He, B.~Yang, and C.~S. Jensen, ``Autocts: Automated
  correlated time series forecasting,'' \emph{PVLDB}, vol.~15, no.~4, pp.
  971--983, 2021.

\bibitem{wu2023autocts+}
X.~Wu, D.~Zhang, M.~Zhang, C.~Guo, B.~Yang, and C.~S. Jensen, ``Autocts+: Joint
  neural architecture and hyperparameter search for correlated time series
  forecasting,'' \emph{SIGMOD}, vol.~1, no.~1, pp. 1--26, 2023.

\bibitem{liang2023mmmlp}
J.~Liang, X.~Zhao, M.~Li, Z.~Zhang, W.~Wang, H.~Liu, and Z.~Liu, ``Mmmlp:
  Multi-modal multilayer perceptron for sequential recommendations,'' in
  \emph{WWW}, 2023, pp. 1109--1117.

\bibitem{li2023automlp}
M.~Li, Z.~Zhang, X.~Zhao, W.~Wang, M.~Zhao, R.~Wu, and R.~Guo, ``Automlp:
  Automated mlp for sequential recommendations,'' in \emph{WWW}, 2023, pp.
  1190--1198.

\bibitem{pan2009survey}
S.~J. Pan and Q.~Yang, ``A survey on transfer learning,'' \emph{TKDE}, vol.~22,
  no.~10, pp. 1345--1359, 2009.

\bibitem{r40}
S.~Hoteit, S.~Secci, S.~Sobolevsky, C.~Ratti, and G.~Pujolle, ``Estimating
  human trajectories and hotspots through mobile phone data,'' \emph{Comput.
  Netw.}, vol.~64, pp. 296--307, 2014.

\bibitem{r30}
Y.~Chen, H.~Zhang, W.~Sun, and B.~Zheng, ``Rntrajrec: Road network enhanced
  trajectory recovery with spatial-temporal transformer,''
  \emph{arXiv:2211.13234}, 2022.

\bibitem{chen2018federated}
F.~Chen, M.~Luo, Z.~Dong, Z.~Li, and X.~He, ``Federated meta-learning with fast
  convergence and efficient communication,'' \emph{arXiv preprint
  arXiv:1802.07876}, 2018.

\bibitem{feng2020pmf}
J.~Feng, C.~Rong, F.~Sun, D.~Guo, and Y.~Li, ``Pmf: A privacy-preserving human
  mobility prediction framework via federated learning,'' \emph{IMWUT}, vol.~4,
  no.~1, pp. 1--21, 2020.

\bibitem{zhang2023simidtr}
Y.~Zhang, L.~Deng, Y.~Zhao, J.~Chen, J.~Xie, and K.~Zheng, ``Simidtr: Deep
  trajectory recovery with enhanced trajectory similarity,'' in \emph{DASFAA},
  2023, pp. 431--447.

\bibitem{chen2023teri}
Y.~Chen, G.~Cong, and C.~Anda, ``Teri: An effective framework for trajectory
  recovery with irregular time intervals,'' \emph{PVLDB}, vol.~17, no.~3, pp.
  414--426, 2024.

\bibitem{r27}
X.~Luo, Y.~Wu, X.~Xiao, and B.~C. Ooi, ``Feature inference attack on model
  predictions in vertical federated learning,'' in \emph{ICDE}, 2021, pp.
  181--192.

\bibitem{r29}
Y.~Liu, Y.~Kang, C.~Xing, T.~Chen, and Q.~Yang, ``A secure federated transfer
  learning framework,'' \emph{IEEE INTELL SYST}, vol.~35, no.~4, pp. 70--82,
  2020.

\bibitem{r10}
L.~Gao, H.~Fu, L.~Li, Y.~Chen, M.~Xu, and C.-Z. Xu, ``Feddc: Federated learning
  with non-iid data via local drift decoupling and correction,'' in
  \emph{CVPR}, 2022, pp. 10\,112--10\,121.

\bibitem{r11}
E.~Diao, J.~Ding, and V.~Tarokh, ``Heterofl: Computation and communication
  efficient federated learning for heterogeneous clients,''
  \emph{arXiv:2010.01264}, 2020.

\bibitem{r12}
Y.~Shi, Y.~Tong, Y.~Zeng, Z.~Zhou, B.~Ding, and L.~Chen, ``Efficient
  approximate range aggregation over large-scale spatial data federation,''
  \emph{TKDE}, vol.~35, no.~1, pp. 418--430, 2021.

\bibitem{r41}
Y.~Liu, J.~J.~Q. Yu, J.~Kang, D.~Niyato, and S.~Zhang, ``Privacy-preserving
  traffic flow prediction: A federated learning approach,'' \emph{IoTJ},
  vol.~7, no.~8, pp. 7751--7763, 2020.

\bibitem{r44}
F.~Yin, Z.~Lin, Q.~Kong, Y.~Xu, D.~Li, S.~Theodoridis, and S.~R. Cui, ``Fedloc:
  Federated learning framework for data-driven cooperative localization and
  location data processing,'' \emph{IEEE Open Journal of Signal Processing},
  vol.~1, pp. 187--215, 2020.

\end{thebibliography}

\end{document}